\pdfoutput=1

\documentclass[11pt]{article}

\usepackage{EMNLP2022}
\usepackage{times}
\usepackage{latexsym}

\usepackage[detect-all]{siunitx}
\usepackage[T1]{fontenc}
\usepackage{ragged2e}
\usepackage{siunitx}
\usepackage{adjustbox}
\usepackage{amsmath,amsfonts,bm}
\usepackage{mathtools}
\newcommand*\rot{\rotatebox{90}}
\newcommand{\STAB}[1]{\begin{tabular}{@{}c@{}}\rot{#1}\end{tabular}}

\usepackage{titlesec}
\titlespacing*{\section}{0.5ex}{1ex}{1ex}
\titlespacing*{\subsection}{1ex}{1ex}{.5ex}
\titlespacing*{\subsubsection}{0pt}{.2ex}{.2ex}

\usepackage{amsfonts}

\usepackage{microtype}

\usepackage[subtle]{savetrees}
\usepackage{multirow}
\usepackage{hyperref}
\usepackage{booktabs} 
\usepackage{tabularx}
\usepackage{graphicx}
\usepackage{subcaption}
\usepackage[htt]{hyphenat}
\usepackage{sidecap}

\usepackage[capitalize,noabbrev]{cleveref}

\usepackage[T1]{fontenc}

\usepackage[utf8]{inputenc}

\usepackage{microtype}

\usepackage{inconsolata}

\title{
Memorization in NLP Fine-tuning Methods
}

\author{Fatemehsadat Mireshghallah\textsuperscript{\rm 1}\thanks{\,\, Corresponding author email: fatemeh@ucsd.edu}, Archit Uniyal\textsuperscript{\rm 2}, Tianhao Wang\textsuperscript{\rm 2},\\
    \textbf{David Evans}\textsuperscript{\rm 2}, \textbf{Taylor Berg-Kirkpatrick}\textsuperscript{\rm 1}\\
    \textsuperscript{\rm 1} University of California San Diego,
    \textsuperscript{\rm 2} University of Virginia \\
    \texttt{[fatemeh, tberg]@ucsd.edu},\\ \texttt{ [a.uniyal,tianhao,evans]@virginia.edu}\\
    }

\begin{document}

\renewcommand{\sectionautorefname}{Section}
\renewcommand{\subsectionautorefname}{Section}
\renewcommand{\subsubsectionautorefname}{Section}

\maketitle

\begin{abstract}

Several recent works have shown that large language models present privacy risks through memorization of training data.
%
%
Little attention, however, has been given to the fine-tuning phase and it is not well understood how memorization risk varies across different fine-tuning methods (such as fine-tuning the full model, the model head, and adapter).
This presents increasing concern as the ``pre-train and fine-tune'' paradigm proliferates.
%
We empirically study memorization of fine-tuning methods using membership inference and extraction attacks, and show that their susceptibility to attacks is very different. We observe that fine-tuning the head of the model has the highest susceptibility to attacks, whereas fine-tuning smaller adapters appears to be less vulnerable to known extraction attacks. 

\end{abstract}


\section{Introduction}

Transformer-based language models have become the models of choice for many NLP tasks, such as email, text and code auto-completion, question answering and sentiment analysis~\cite{chen2021evaluating,dai-smartcompose-2019}. 
These models are commonly trained using the {\it pre-train and fine-tune paradigm}, where they are first trained (\emph{pre-trained}) on a  large, general domain dataset (in the order of hundreds of Gigabytes), and then \emph{fine-tuned} on smaller, task-specific datasets to adapt the model to a specific domain~\cite{ramponi2020neural,li2021prefix,houlsby2019parameter}. 

Several works have demonstrated that such large models have a high capacity for memorizing training samples during pre-training and are therefore highly susceptible to membership inference and data extraction attacks~\cite{  10.1145/3372297.3417880, https://doi.org/10.48550/arxiv.2012.07805, https://doi.org/10.48550/arxiv.2101.00036}.
More specifically, \citet{https://doi.org/10.48550/arxiv.2012.07805} and \citet{mireshghallah2022quantifying} have mounted such attacks on pre-trained language models and shown the severity of this issue by extracting complete training sequences and inferring membership of a large fraction of the training samples. 

These works focused on memorization during pre-training, but scant attention has been given to fine-tuning.  In this work, we focus on different fine-tuning methods and their propensity for memorization of training samples. 
Fine-tuning data is actually of higher concern than pre-training data, since most pre-training datasets are large public corpora~\cite{raffel2019exploring,Dodge2021DocumentingTE} with limited privacy concerns~\cite{brown2022does}, while fine-tuning sets are small, targeted, and potentially very private~\cite{basu2021benchmarking,li2021large}.
%
Further, pre-training generally happens only a few times (as it needs resources that are usually only available to large companies~\cite{brown2020language}) while fine-tuning is increasingly the dominant way that end-users fit models.

\begin{figure}[t]
    \centering
     \includegraphics[width=1.0\linewidth]{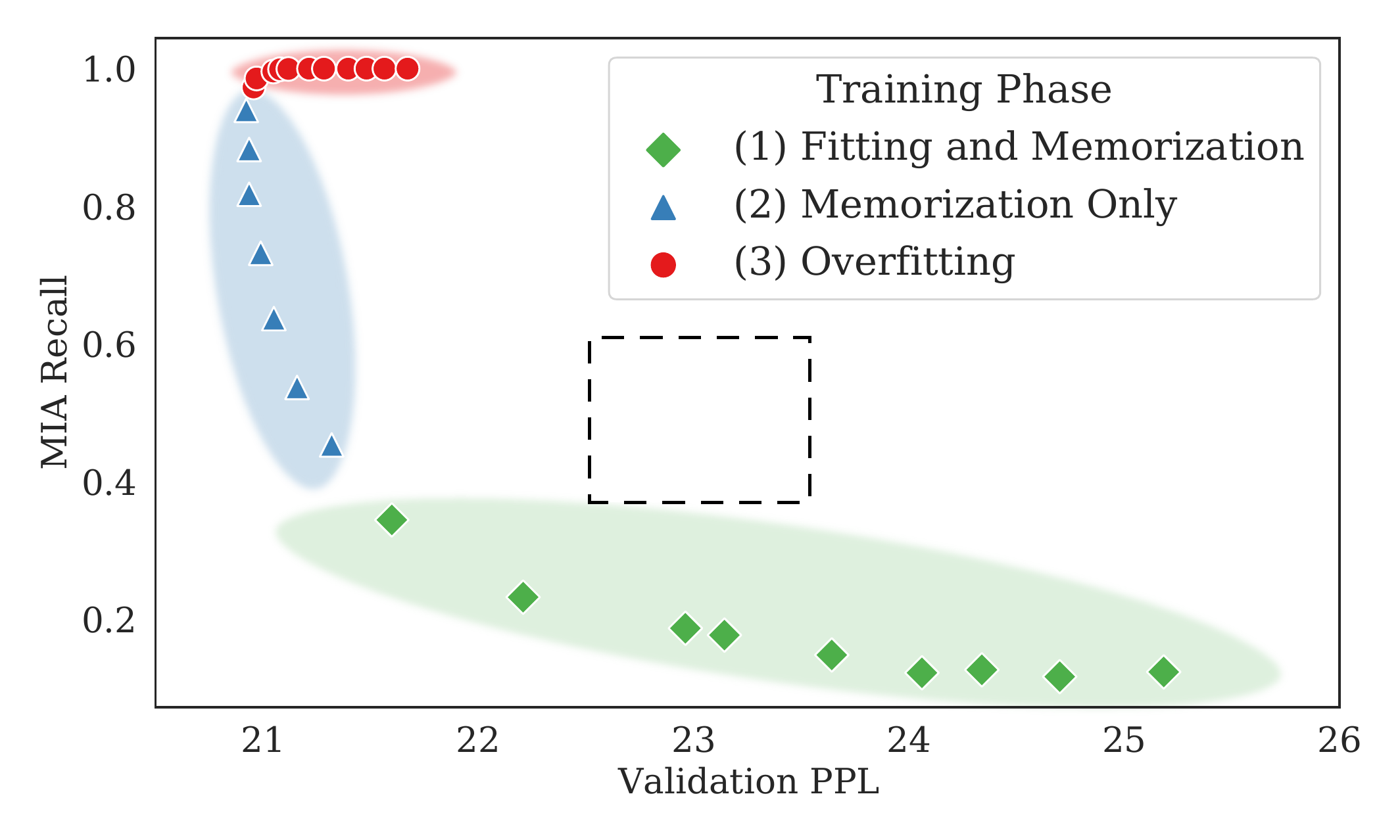}
     \footnotesize
     \caption{Each point in the graph shows the given metric values at the end of each training epoch. The rightmost lower points show the beginning, and as we move to left and upwards training progresses. We identify three separate phases within the learning process, distinguished by their memorization and generalization trends.}
    \label{fig:overview}
    \vspace{-3ex}
\end{figure}
Given the size of these large language models, fine-tuning all the model parameters can be compute and memory-intensive~\cite{ lewis2019bart, brown2020language, fedus2021switch}.
As a result, recent works have proposed new parameter efficient fine-tuning methods that update only a subset of the model's parameters~\cite{houlsby2019parameter, li2021prefix, https://doi.org/10.48550/arxiv.2110.04366}. 
In this paper, we focus on studying memorization of three popular fine-tuning methods: 
(1) fine-tuning all model parameters (2) fine-tuning the head, which is commonly used by practitioners and involves updating only the last layer of the model which produces the logits, and (3) fine-tuning adapters~\citep{houlsby2019parameter},  which are small bottleneck modules inserted within transformer blocks. 
For measuring memorization, we use two proxy metrics: (a) recall of a reference-based membership inference attack (MIA)~\cite{mireshghallah2022quantifying} and (b) exposure~\cite{carlini2019secret}, which measures how susceptible the model is to a sample extraction attack which tries to reconstruct samples from training data.
We run our experiments on the Wikipedia~\cite{merity2016pointer}, Penn Treebank~\cite{marcus-etal-1993-building} and Enron Emails~\cite{klimt2004enron} datasets, for the task of autoregressive language modeling. We selected Wikipedia and Penn Treebank as they are most commonly used for fine-tuning, and  Enron since it is a dataset of emails representing  private tuning data.
Figure~\ref{fig:overview} shows how we conceptually identify three distinct phases in the fine-tuning process, based on validation perplexity (generalization) and membership inference attack recall (memorization). Each point shows these metrics at the end of a training epoch.
For all fine-tuning methods, we observe that in a~\emph{memorization only} phase, the model memorizes more and more, without overfitting or generalizing better (Figure~\ref{fig:pareto}). 
In terms of different fine-tuning methods, we find that the common practice of fine-tuning only the head of a model has the highest memorization (by a large margin) for the same level of perplexity, among different fine-tuning methods -- even full fine-tuning, which updates more parameters.
This result is surprising and potentially indicates that only tuning parameters higher in the model architecture (closer to the output) exacerbates the memorization and increases the leakage based on our metrics.
We also show that fine-tuning the full model and small adapters are on the Pareto-frontier in terms of the attack recall vs.\ validation perplexity graph. 
Code and instructions to reproduce our results are available at~\url{https://github.com/mireshghallah/ft-memorization/}. 
\section{Model Fine-tuning}

%
%
%
%
We focus on two main fine tuning methods, for fine-tuning GPT-2 with next word prediction objective: (1) fine-tuning the model head, i.e., the prediction layer, as it is the most common method used in practice, and (2) fine-tuning adapters~\cite{houlsby2019parameter}. Adapters are small rank-restricted modules that are inserted inside transformer blocks, as added parameters and are fine-tuned for different tasks or datasets. 
The shape and size of the adapter module is controlled by the \emph{reduction factor}, which determines the ratio of the size of the bottleneck to its input. %
During adapter tuning, the rest of the model remains frozen, therefore the number of trainable parameters is low (around $1\%$ of the full model parameters).  
%
%
In our experiments, we choose reduction factors of $16$ and $2$, for adapters, as the former is the default used by~\cite{pfeiffer2020AdapterHub,houlsby2019parameter}, and the latter is the largest factor.
%

\section{Measuring Memorization} \label{sec:mem}

To measure memorization, we use two metrics: membership inference attack recall and exposure. 

\noindent\textbf{Membership Inference (MIA Recall).} We use the percentage of training samples that are correctly classified as training members (out of a pool of training and validation samples) by the reference-based attack proposed in~\citet{mireshghallah2022quantifying} and \citet{carlini2021membership} as a proxy metric of memorization.  
For each sample $x$ whose membership in the training set we want to determine, we feed it to the fine-tuned model, $M$, and get its likelihood, $\textbf{Pr}^M(x)$. We also feed it to a reference model, $R$, a pre-trained model that is not fine-tuned, and get the probability $\textbf{Pr}^R(x)$.
We then use $\mathit{LR}(x)=\frac{\textbf{Pr}^R(x)}{\textbf{Pr}^M(x)}$, the likelihood ratio, to determine if $x$ is a training sample. If $\mathit{LR}(x)$ is smaller than threshold $t$, we classify it as a training set member. 
Otherwise, we classify it as a non-member. We determine the threshold $t$ by calculating $\mathit{LR}(s)$ for all $s$ in the validation set, and then choose the threshold to be the highest threshold such that the false positive rate (over training and validation members) would not exceed $10\%$. 
%
The higher the recall of this attack is, the higher the leakage of the model. 
%

\noindent\textbf{Exposure.\ } 
As a second measure of memorization, we use the exposure metric from \citet{carlini2019secret} which inserts a secret (canary) of a certain format into the training data and calculates its vulnerability to extraction. Exposure is defined as the negative log-rank  of the inserted secret in terms of model probability, among all other possible sequences of the same length. This quantity is then added to a constant to ensure the exposure is always positive. The lower the exposure is, the harder it is to extract the secret. In our experiments, we insert 50 copies of the phrase ``the secret number is 940955'' into the training data to accentuate the differences between the fine-tuning methods. For a six-digit secret, an exposure of around $\log_2(10^6)\approx 20$ means the canary can be reliably extracted from the model.
\begin{figure}[]
    \centering
    
    \begin{subfigure}{0.48\textwidth}
     \includegraphics[width=\linewidth]{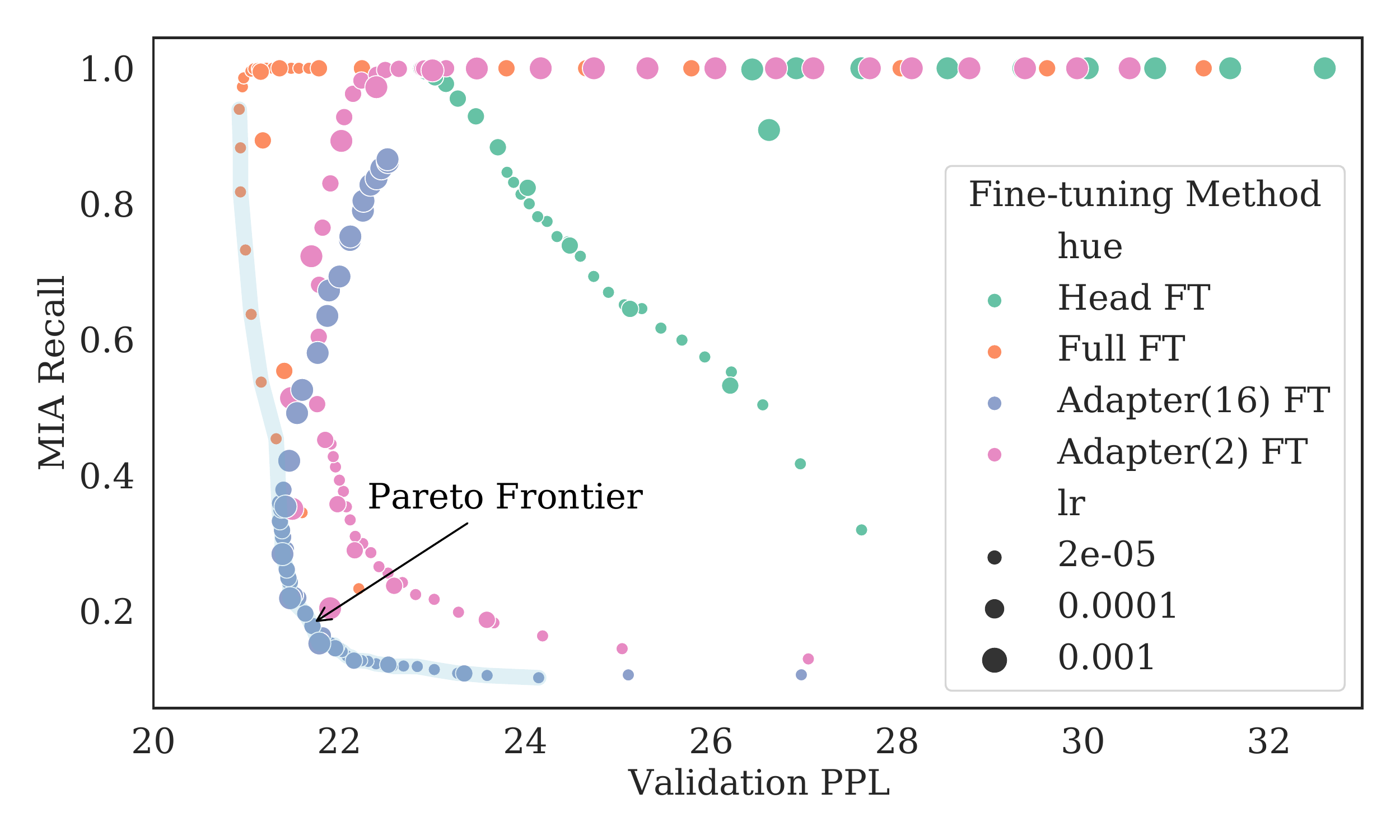}
     \footnotesize
     \caption{Wikipedia Dataset}
     \label{fig:pareto:wikipedia}
    \end{subfigure}
    \begin{subfigure}{0.48\textwidth}
     \includegraphics[width=\linewidth]{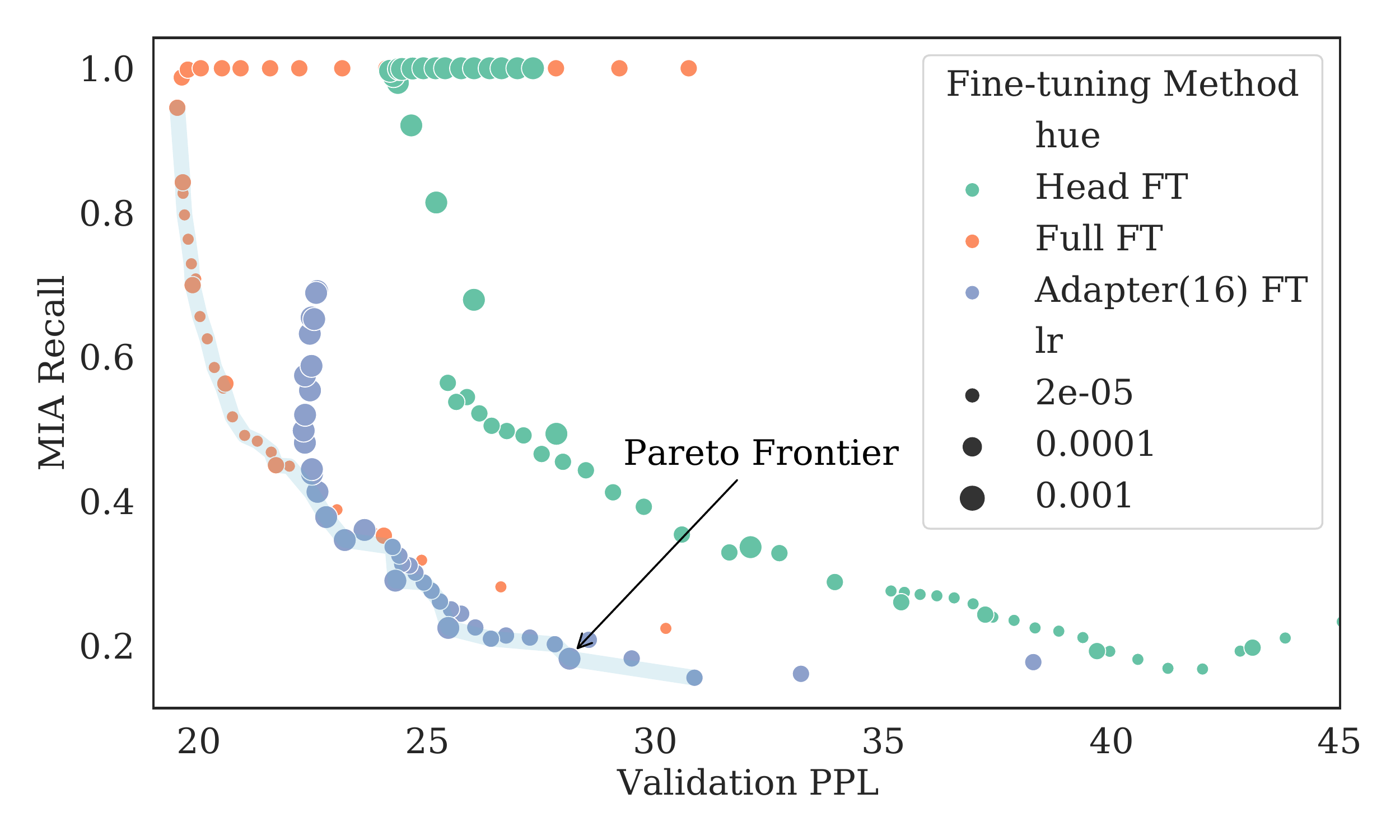}
    \footnotesize 
     \caption{Penn Treebank Dataset}
     \label{fig:pareto:ptb}
    \end{subfigure}
    \begin{subfigure}{0.48\textwidth}
     \includegraphics[width=\linewidth]{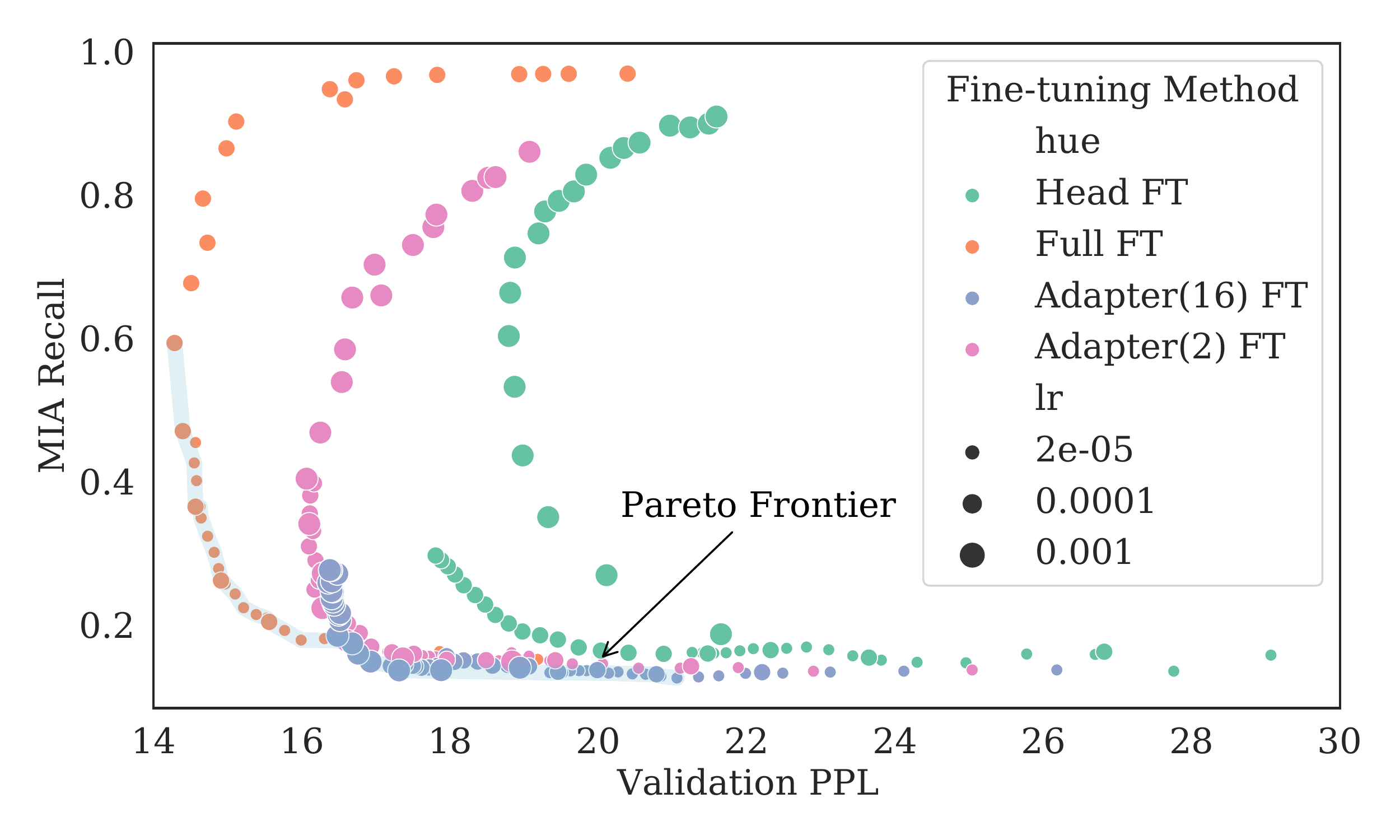}
     \footnotesize
     \caption{Enron Dataset}
     \label{fig:pareto:enron}
    \end{subfigure}
  \caption{Pareto frontier for utility (validation PPL) Vs. privacy (MIA recall). Each dot shows different checkpoints, and the colors show different fine-tuning methods. We desire models that have low PPL and low attack recall.  } 
    \label{fig:pareto}
    \vspace{-2ex}
\end{figure}

\begin{figure}[t]
    \centering
     \includegraphics[width=0.96\linewidth]{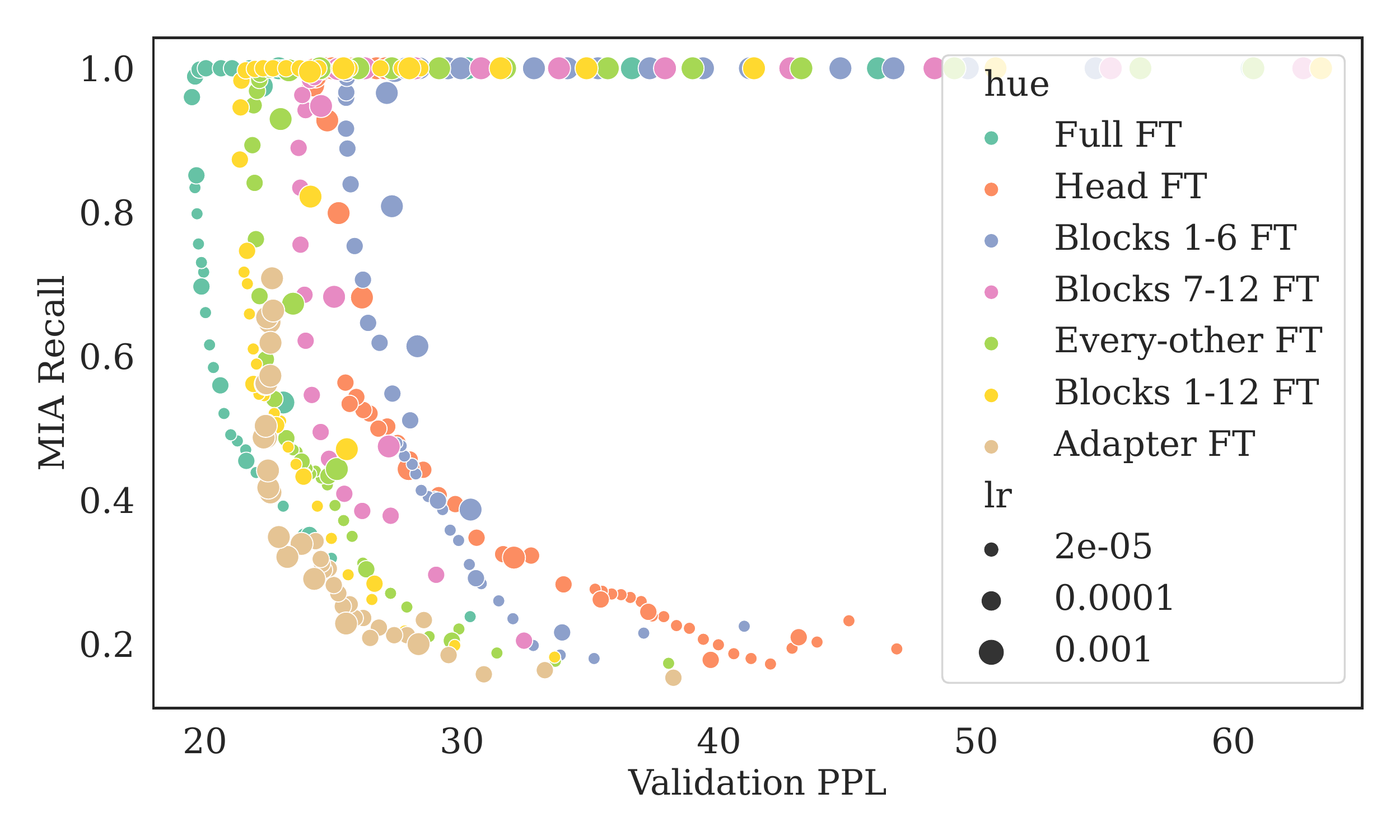}
     \footnotesize
     \caption{Ablating how the location and number of trainable parameters effects memorization on the Penn Treebank dataset. Each dot shows different checkpoints, and the colors show different fine-tuning methods. We desire models that have low PPL and low attack recall.}
    \label{fig:ablation}
    \vspace{-3ex}
\end{figure}

\section{Experimental Setup}

\noindent\textbf{Datasets.}
(1) Huggingface's Wikipedia \textsf{wikitext-2-raw-v1} dataset, consisting of $36718$ training samples (2) Huggingface's Penn Treebank \textsf{ptb\_text\_only}, consisting of $42068$ training samples and (3) a sub-sampled version of Enron email dataset consisting of $7180$ emails. We use a sequence length of $1024$, training batch size of $8$, and fine-tune for 20 epochs. %

\noindent\textbf{Models.}
We study memorization in fine-tuning Huggingface's pre-trained GPT-2 on the datasets mentioned above.
We use a pre-trained but not fine-tuned GPT-2 as the reference model for our membership inference attack.
We use the adapter hub's implementation of the Pfeiffer architecture, with reduction factors $2$ and $16$~\cite{pfeiffer2020AdapterHub}.

\noindent\textbf{Metrics.}
We use \textit{Validation Perplexity} as a metric for the performance of the model, where lower perplexity is better. 
We evaluate memorization at each epoch using the ~\textit{MIA recall} and \textit{exposure} metrics described in \autoref{sec:mem}. 
The experiments in the paper are all repeated 3 times and we report the average values for each metric. 

\noindent\textbf{Hyperparameters and result presentation.} We run optimization for each fine-tuning method for $20$ epochs, and perform evaluation of the mentioned metrics at the end of each epoch. We  experiment with the three learning rates $2\times 10^{-5}, 10^{-4}, 10^{-3}$, and present the results for all of them. 
%
Therefore, each graph would have an overall of $20\times3$ points, for each fine-tuning method, unless the point is outside the plot range.
For the reported exposure numbers, we  selected points close to the pareto frontier to present in Table~\ref{tab:exposure}, to summarize results.
%
%

\begin{table}[]
    \centering
    \caption{Exposure metric. \small \rm 
    Higher exposure indicates more leakage, and exposure above $20$ means the secrets (canaries) are reliably extractable. The perplexity numbers here are different from the ones in other experiments since the training data is diluted with the artificially inserted secrets.}
    \vspace{-2ex}
    \label{tab:exposure}
    \begin{adjustbox}{width=0.92\linewidth, center}

\begin{tabular}{@{}clllll@{}}
	\toprule
&   & {Full FT} & {Head FT} & {Adapters (2)} & {Adapters (16)}  \\
    \midrule
\multicolumn{2}{c}{Parameters (Millions)} &	124.440 &	38.590 & 7.092 &	0.895   \\
\midrule

    \multirow{2}{*}{\STAB{Wiki}} 
&Val PPL   &    \textbf{24.82}   &	28.76      &	24.41  &	25.26	\\
&Exposure  & 	1.42	&   10.78      &	14.54  &	\textbf{0.83} \\
\midrule
    \multirow{2}{*}{\STAB{PTB}} 
&Val PPL    &  29.55   &	31.24    &	29.79  &	\textbf{29.41}  	\\
&Exposure	&   7.03	&   12.0    &	12.40  &	\textbf{4.54}   \\
\midrule
    \multirow{2}{*}{\STAB{Enron}} 
&Val PPL    &   \textbf{12.52}   &	13.51    &	13.03  &	12.81  	\\
&Exposure	&   1.32	&   10.77     &	2.02  &	\textbf{0.440}  \\
	\bottomrule
\end{tabular}

    \end{adjustbox}
    \vspace{-2ex}
\end{table}
\begin{table*}[t]
    \centering
    \caption{Comparison of fine-tuning different transformer blocks on the Wikipedia dataset.}
    \vspace{-2ex}
    \label{tab:sample-out-model}
    \begin{adjustbox}{width=0.85\linewidth, center}

\begin{tabular}{@{}llllllllll@{}}
	\toprule
	&   &  {Block 1} & {Block 5} &  {Block 8} &{Block 12} & {Full FT} & {Head FT} & {Adapters (2)} & {Adapters (16)}   \\
\midrule[0.1pt]
\multirow{3}{*}{}
& Validation PPL  &   24.39  &	23.35   &	23.36	   & 24.05  &	23.05   &	23.93   &	23.62 & \textbf{21.75} 	   \\
&MIA Recall  &22.2 &	22.6 &	20.8	&   21.3   &	19.2   &	81.6   &	16.8  & \textbf{15.2}   \\
&\#Params (in Millions) & 7.088 &	7.088   &	7.088   &	7.088   &	124.440 &	38.590 & 7.092 &	0.895   \\
	\bottomrule
\end{tabular}

    \end{adjustbox}
\end{table*}

\section{Results}

In this section we discuss our experimental results comparing the privacy-utility trends for different fine-tuning methods. 
%
%
We refer to the naming convention shown in \autoref{fig:overview}
and provide extended graphs for each experiment in Appendix~\ref{app:separate}. 
We also present additional experiments where we train the model from scratch (instead of fine-tuning pre-trained models),  fine-tune different model architectures, and study the generalization gap   in Appendix~\ref{app:additional}.
%



\subsection{Memorization of Fine-tuning Methods}

Figures~\ref{fig:pareto:wikipedia},~\ref{fig:pareto:ptb},~\ref{fig:pareto:enron} compare the fine-tuning methods in terms of privacy leakage, measured by MIA recall and Table~\ref{tab:exposure} shows the exposure results for the three datasets, along with their parameter counts.
The blue lines show the Pareto frontier, marking the desirable trade-off points, with low recall and  PPL.

\subsubsection{Shared  Trends}
The ``memorization only'' phase in training, where validation perplexity (generalization) is stable and the model has not yet overfit, is also observed by~\citet{tanzer-etal-2022-memorisation} in pre-trained BERT-based classifiers. However, it is named the ``settling phase'' there, and it is suggested that as \emph{validation perplexity} is rather stable, early stopping is not important and training can stop at any point before overfitting. We, however, show that \emph{memorization} is actually increasing during that phase. Therefore, if we are optimizing for privacy as well, it is best to stop training earlier. 
Appendix~\ref{app:gen-valppl} shows generalization gap vs.\ validation perplexity graphs demonstrating that the gap remains stable during the ``memorization only'' phase.
For all the methods, across all datasets, in the ``fitting+memorization'' and the ``memorization only'' phases, we see an increase in memorization, without any overfitting. This shows that we can have high memorization/learning, and still not overfit.
This is also observed for training large language models from scratch in
~\citet{tirumala2022memorization}, which focuses on analyzing the effect that text type (e.g., part of speech, numbers), data size and model size have on memorization when training from scratch. 
%

\subsubsection{Comparison of Fine-tuning Methods}\label{sec:comp}
Results for both the MIA recall and exposure metrics (\autoref{fig:pareto} and \autoref{tab:exposure}) are consistent, showing higher leakage for head fine-tuning and lower for full model fine-tuning and adapters. 
%
%
%
The first observation here is that head fine-tuning is an outlier, with extremely high leakage, on all three datasets. We can also see that the validation perplexity achieved by this method is consistently lower than the other methods.
%
We hypothesize that the high leakage of fine-tuning the head is due to both the high number of parameters ($38$ million) and the location of the parameters, right at the last layer of the model where the next word prediction happens. 
While full fine-tuning actually touches more parameters than head fine-tuning, it leads to less leakage under the attacks we investigate. This result is somewhat surprising and potentially indicates that tuning parameters lower in the model architecture mitigates some of the explicit memorization performed by the head. We further study this phenomenon and ablate it in Section~\ref{untie}.

We also observe that for a low-perplexity regime (without considering the cost), full fine-tuning is the best choice as it offers utility superior to adapters.
%
However, if we have tolerance for higher perplexity, to get lower leakage, opting for adapters with a reduction factor of 16 appears better as it has lower MIA recall and a lower propensity for overfitting, compared to the other methods. 
One final observation is that full-finetuning has the shortest ``fitting+memorization'' phase, whereas head fine-tuning has the longest.


\subsection{Parameter Count, Location and Tying}\label{untie}
To further test our  hypothesis that the privacy-utility trade-off has to do with \textit{both} trainable parameter count and location/distribution within the model architecture (\autoref{sec:comp}), we run experiments with the following set of trainable parameters: (1)~first half: blocks 1--6 of the 12 transformer blocks of the GPT2 model (42M trainable params), (2)~second half: blocks 7--12 (42M), (3)~every other block (42M) and (4)~entire body: all the 12 blocks (84M ). In all these scenarios we freeze the head and fine-tune only the blocks. As shown in Figure~\ref{fig:ablation}, we find that Full FT~$>$ Adapters~$>$ all 12 blocks$=$every~other~block~$>$ blocks 7 to 12~$>$ blocks 1 to 6 $>$ Head FT, in terms of privacy-utility trade-off desirability.
Based on this, we argue that how the trainable parameters are scattered in the network affects how well the model makes progress in the first phase (the training and fitting phase), which affects the validation perplexity when it enters the second phase (memorization-only phase). As \autoref{fig:pareto} also shows, full fine-tuning and adapter tuning make faster progress and end up in a lower perplexity. 

\begin{figure}[t]
    \centering
     \includegraphics[width=0.96\linewidth]{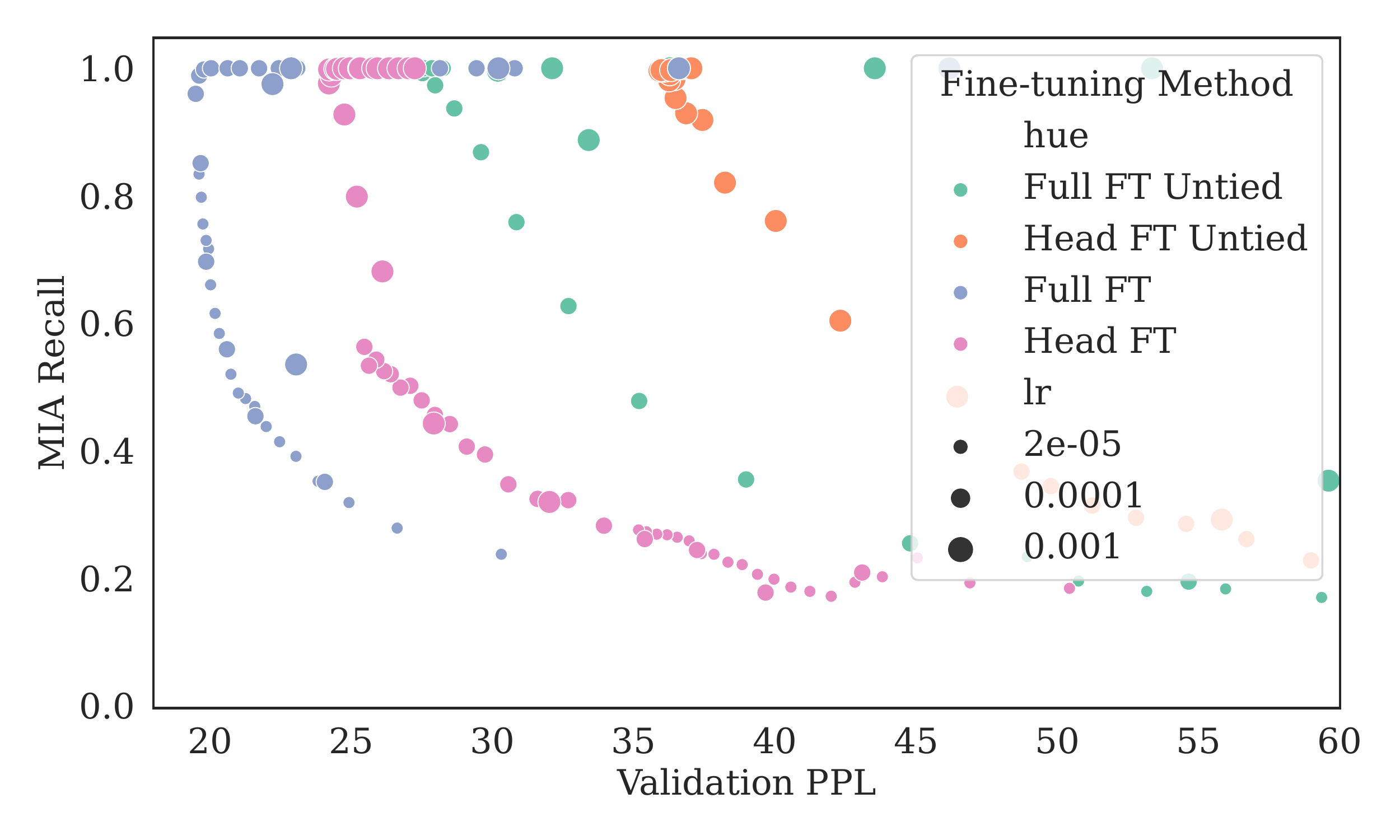}
     \footnotesize
     \caption{Ablating how the untying of the trainable parameters effects memorization on the Penn Treebank dataset. Each dot shows different checkpoints, and the colors show different fine-tuning methods. We desire models that have low PPL and low attack recall.}
    \label{fig:ablation-tie}
    \vspace{-3ex}
\end{figure}


~\autoref{fig:ablation-tie} shows an ablation study of how untying model parameters affects the privacy-utility trade-off. By untying parameters, we mean creating a separate set of parameters for the head of the model and the input embeddings, as by default these two parameter sets are tied in GPT2, meaning the same set of $38.59$ Million parameters are used for both these components. However, in the untied scenario, we first duplicate them, and then create separate trainable parameters, adding an extra set of $38.59$ Million trainable parameters to the model. As the figure shows,  tying the parameters improves the progress in training and puts the model at an advantage, compared to untying them, creating a better overall privacy-utility trade-off.


\subsection{Fine-tuning Single Transformer Blocks}

To have a full analysis of fine-tuning leakage,  we also look at fine-tuning individual adapter blocks and freezing the rest of the model.
The GPT-2 model has 12 blocks, and we experiment with fine-tuning the first, 5th, 8th, and 12th block, to cover different positions within the model. 
\autoref{tab:sample-out-model} shows the results for this experiment. We have selected the numbers such that the validation PPLs are as similar as possible. 
There does not seem to be any significant difference between fine-tuning different blocks, as they all manifest similar attack recalls. 
Block 8's recall, however, is lower than other blocks, with lower PPL, which would make it the most desirable block for fine-tuning in terms of the PPL-leakage trade-off. 
With respect to privacy-utility tradeoffs, fine-tuning full blocks seems less desirable than using adapters or fine-tuning the entire model.


\section{Conclusion}

When fine-tuning is done using sensitive training data, it is important to not just consider the cost and utility of fine-tuning methods but to also be aware that they may have different risks in terms of privacy. Our experiments show that 
%
the common practice of fine-tuning only the head of a model has the highest memorization (by a large margin).
Full model fine-tuning and adapter tuning, however, are both on the Pareto-frontier in terms of attack recall vs.\ validation perplexity, suggesting that they are more suitable when privacy is a concern.
%
%


\section*{Acknowledgements}
This project is funded in part by the NSF under grant 2200333. The authors would like to thank the anonymous reviewers and meta-reviewers for their helpful feedback. We also thank Nikolai Vogler, Nikita Srivatsan, and Kazem Taram for insightful discussions. Additionally, we thank our colleagues at the UCSD Berg Lab and UVA Security Research Group for their helpful comments and feedback. 

\section*{Limitations and Ethics Statement}
In our study we  focus on autoregressive language models -- specifically GPT-2, as it has been shown to be more prone to memorizing samples than pre-trained masked language models (MLM)~\cite{carlini2021extracting,lehman-etal-2021-bert}
Also, in this paper we loosely refer to the recall of the membership inference attack on the training set as memoirzation. However, we need to keep in mind that a low attack recall does not necessarily mean low memorization, and there might be stronger attacks (of other types, such as reconstruction) that can better uncover memorization in language models. 

In this work we have used publicly available datasets and have not collected any sensitive/private data. The ultimate goal of our study is to contribute to analyzing  memorization under different fine-tuning paradigms, thereby advancing our intuition of how we can better deploy private, fair and safe language models.
%


\bibliography{anthology,emnlp2022}
\bibliographystyle{acl_natbib}

\newpage
\appendix
\clearpage
\section{Appendix}

\subsection{Additional Experiments}\label{app:additional}
\subsubsection{Correlation between Generalization and Memorization}\label{app:gen-mia}

Figure~\ref{fig:gap} shows the correlation between the generalization gap and membership inference attack recall. 
The generalization gap refers to the subtraction of training perplexity from validation perplexity, and a larger gap means more overfitting.
We can see that there is a direct relation between the generalization gap and attack recall, for all fine-tuning methods.
We can also see that for Penn Treebank and Enron, head fine-tuning has a consistently higher generalization gap, which could explain why the membership inference attack is more successful on it.
\subsubsection{Generalization Gap vs Val PPL}\label{app:gen-valppl}

Figure~\ref{fig:gap} shows generalization gap (validation$-$train perplexity) versus validation perplexity. We plot this to show how this differes from MIA recall (memorization) versus perplexity (Figure~\ref{fig:pareto}), and to emphasize how in the \emph{memorization only} phase, memorization is increasing (the long vertical stretch in Figure~\ref{fig:pareto}), however the validation perplexity and generalization gap remain almost the same (the sharp turning point in Figure~\ref{fig:gap}).

%

\begin{figure}[t]
    \centering
     \includegraphics[width=0.96\linewidth]{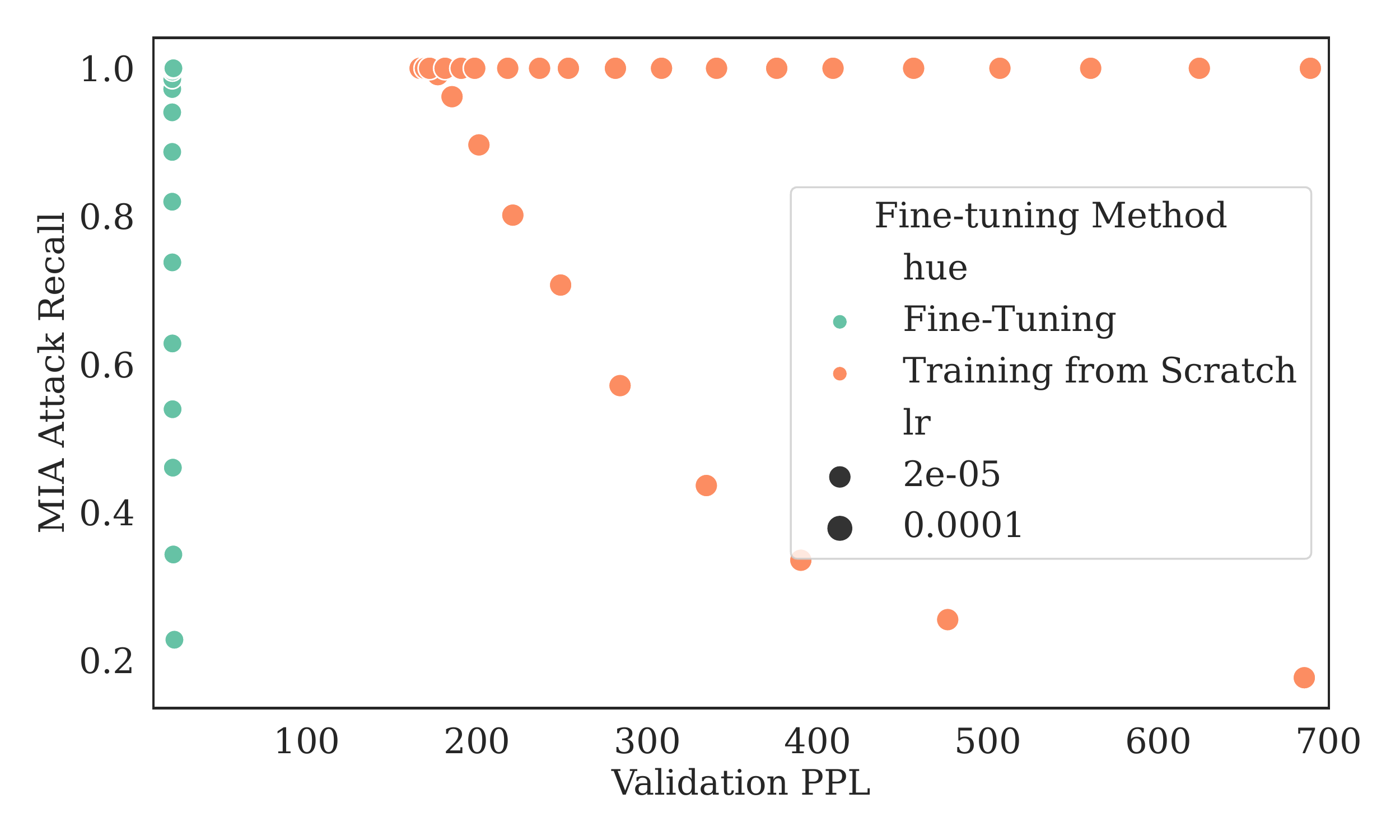}
     \footnotesize
     \caption{Ablating how training the model from scratch affects the privacy-utility trade-off, compared to fine-tuning a pre-trained model, on the Wikipedia dataset. Each dot shows different checkpoints, and the colors show different fine-tuning methods. We desire models that have low PPL and low attack recall.}
    \label{fig:scratch}
    \vspace{-3ex}
\end{figure}

%
\subsection{Training From Scratch}
Figure~\ref{fig:scratch} shows how pre-training a finetuned mdoel is different from training the model from scratch, in terms of validation perplexity and attack recall. We can see that  fine-tuning a pre-trained model leaks less information, than fine-tuning from scratch.

\subsubsection{Other Models}

\begin{figure}[]
    \centering
    
    \begin{subfigure}{0.48\textwidth}
     \includegraphics[width=\linewidth]{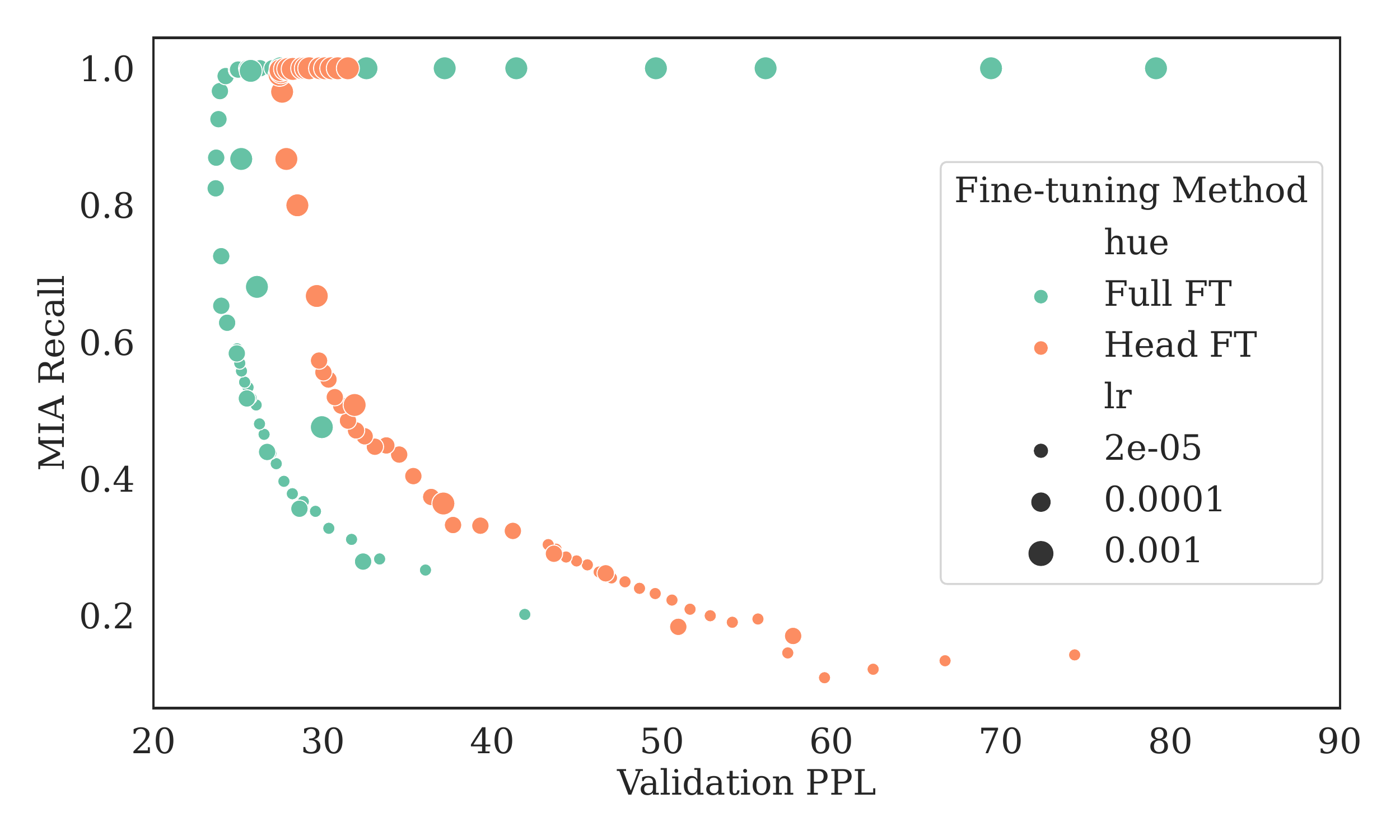}
     \footnotesize
     \caption{DistilGPT2}
     \label{fig:other-model:openai-gpt}
    \end{subfigure}
    \begin{subfigure}{0.48\textwidth}
     \includegraphics[width=\linewidth]{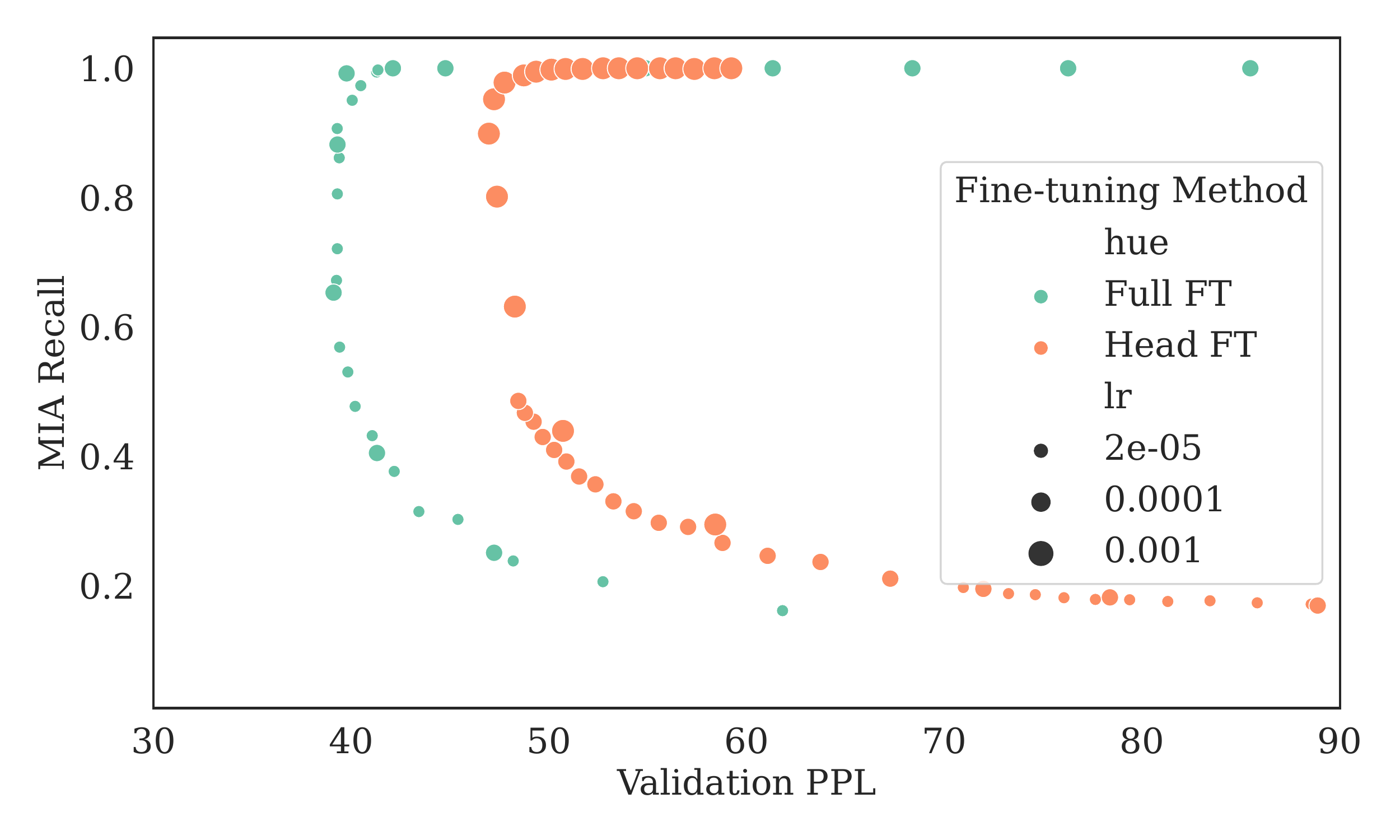}
    \footnotesize 
     \caption{OpenAI-GPT}
     \label{fig:other-model:distilgpt2}
    \end{subfigure}
  \caption{Utility Vs. privacy (MIA recall) on the Penn Treebank dataset for DistilGPT2 and OpenAI-GPT models. Each dot shows different checkpoints, and the colors show different fine-tuning methods. We desire models that have low PPL and low attack recall.  } 
    \label{fig:other-model}
    \vspace{-2ex}
\end{figure}

To further test how our findings generalize to other models, we repeat our experiments on the Huggingface ~\texttt{distilgpt2} and \texttt{openai-gpt} as well, and show the results in Figure~\ref{fig:other-model}. As we see, the results are commensurate with those of GPT2. We cannot run experiments with adapters here as these models are not supported by the adapter library yet. 

\subsection{Separate Plots}~\label{app:separate}
\begin{figure}[]
    \centering
    
    \begin{subfigure}{0.48\textwidth}
     \includegraphics[width=\linewidth]{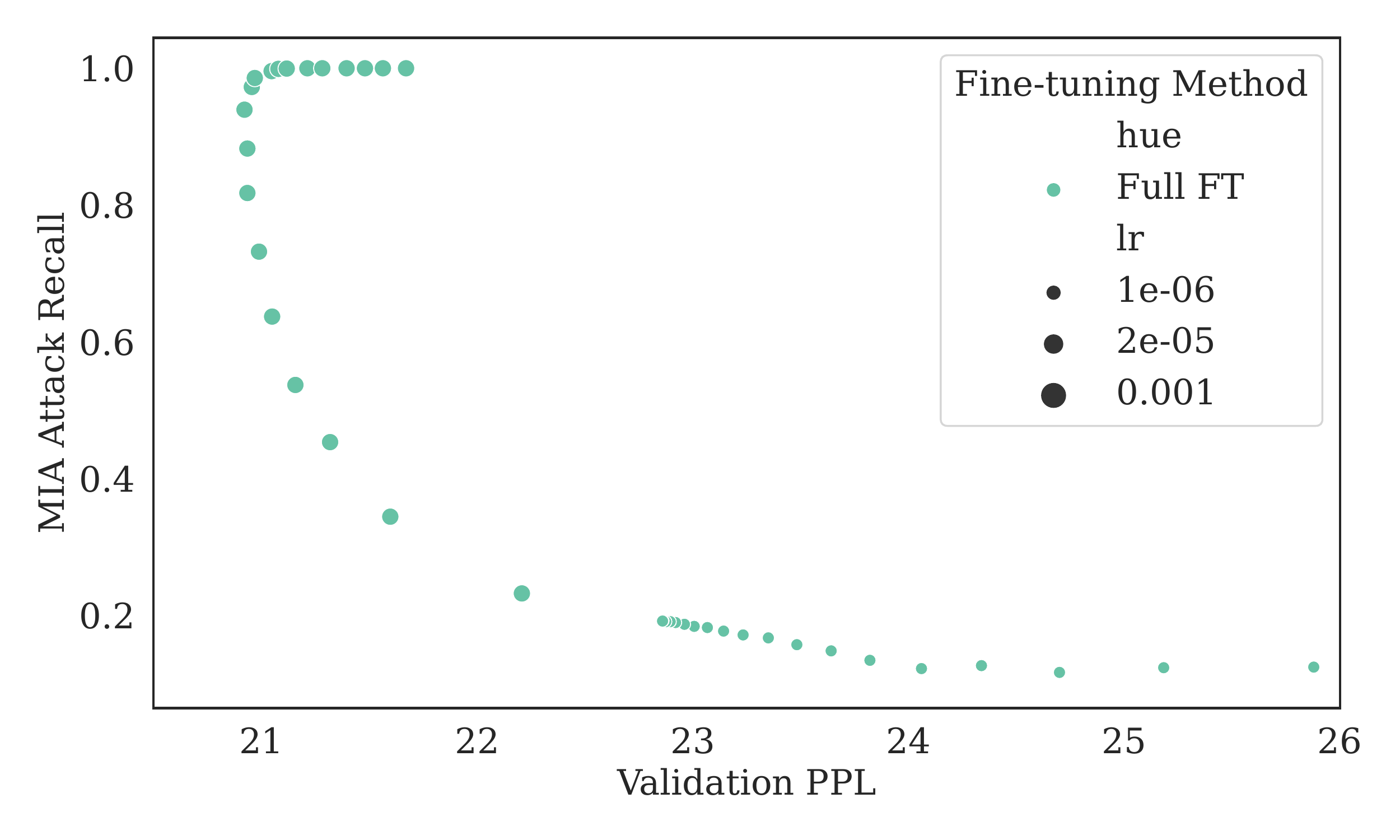}
     \footnotesize
     \caption{Full FT}
     \label{fig:sep:wikipedia:full}
    \end{subfigure}
    \begin{subfigure}{0.48\textwidth}
     \includegraphics[width=\linewidth]{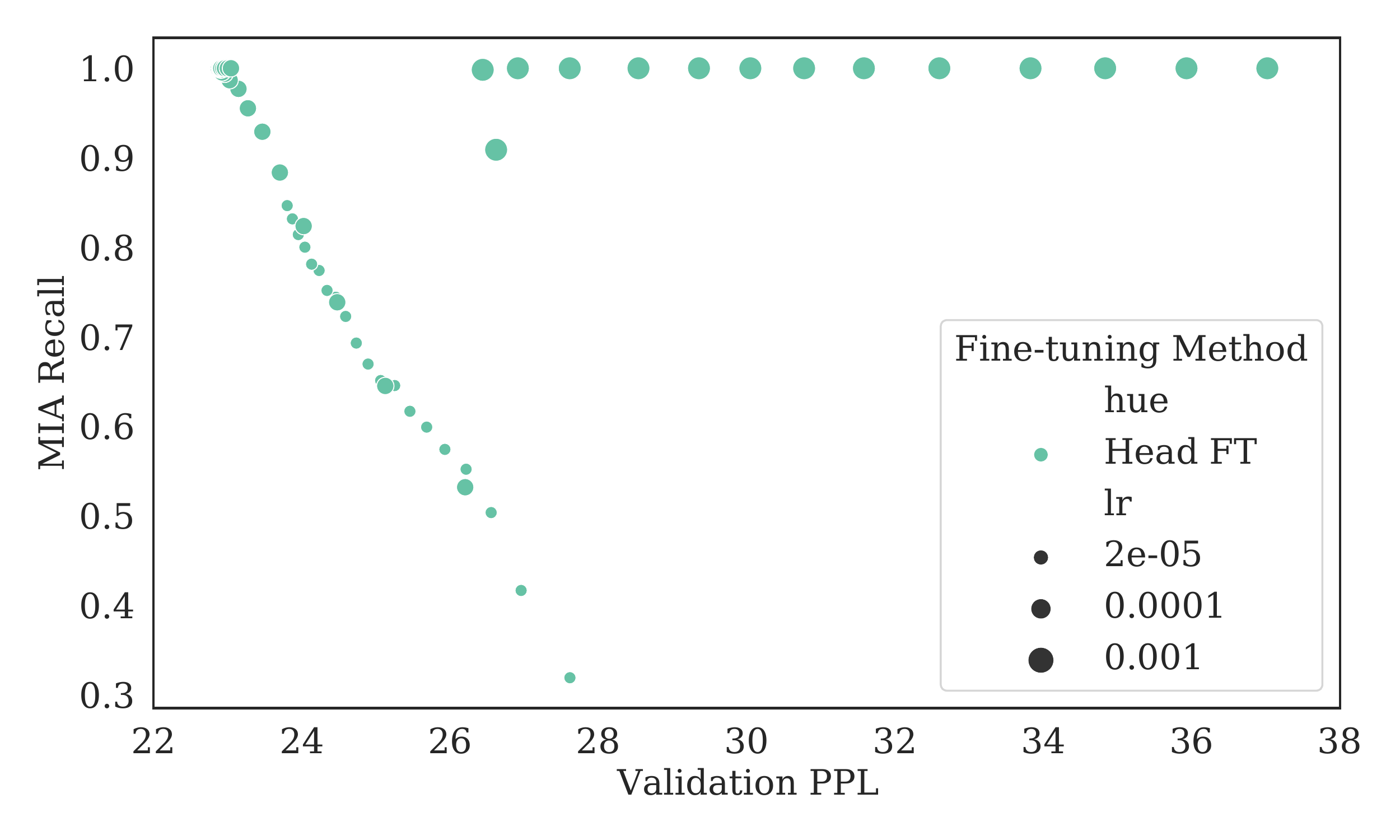}
    \footnotesize 
     \caption{Head FT}
     \label{fig:sep:wikipedia:head}
    \end{subfigure}
    \begin{subfigure}{0.48\textwidth}
     \includegraphics[width=\linewidth]{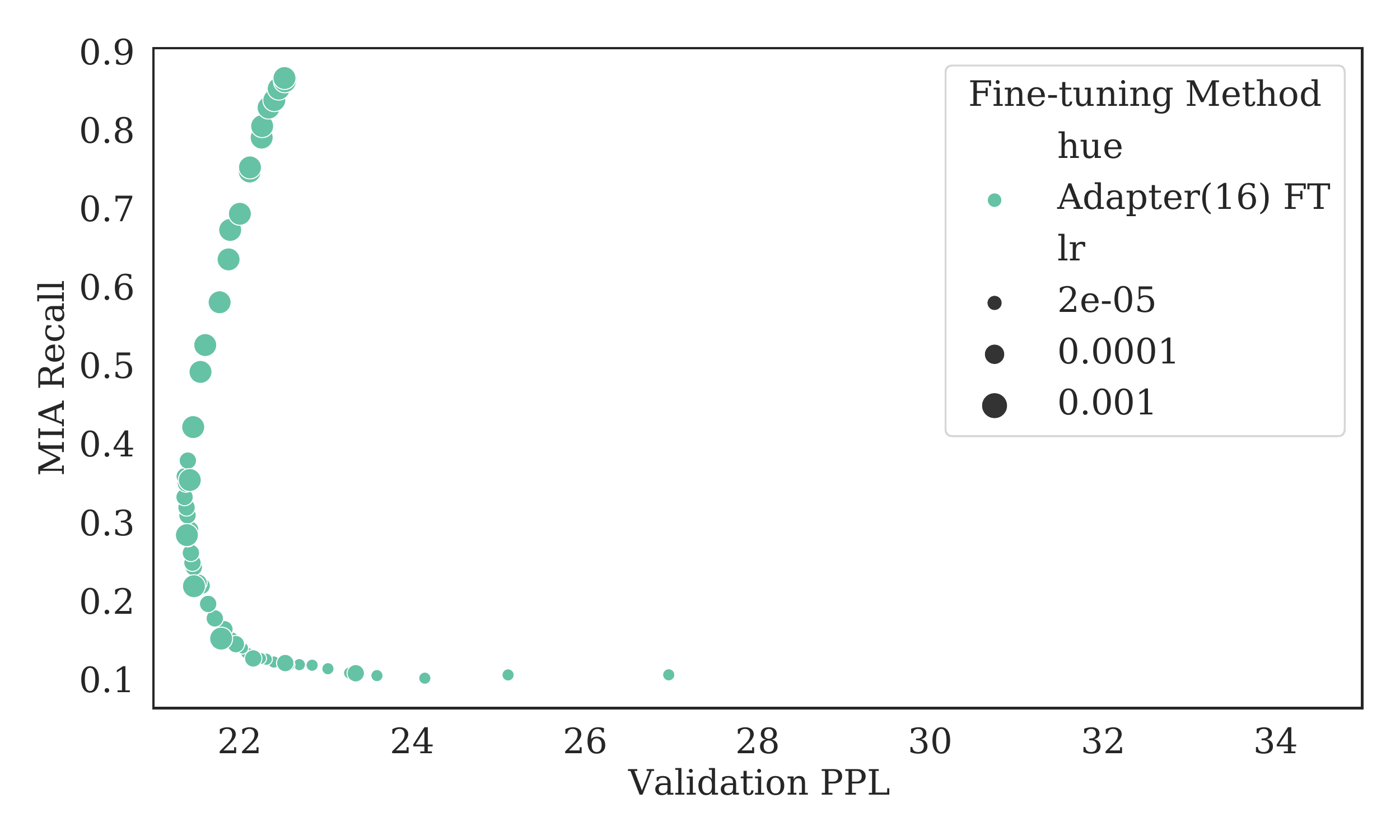}
    \footnotesize 
     \caption{Adapter(16) FT}
     \label{fig:sep:wikipedia:adp16}
    \end{subfigure}
    \begin{subfigure}{0.48\textwidth}
     \includegraphics[width=\linewidth]{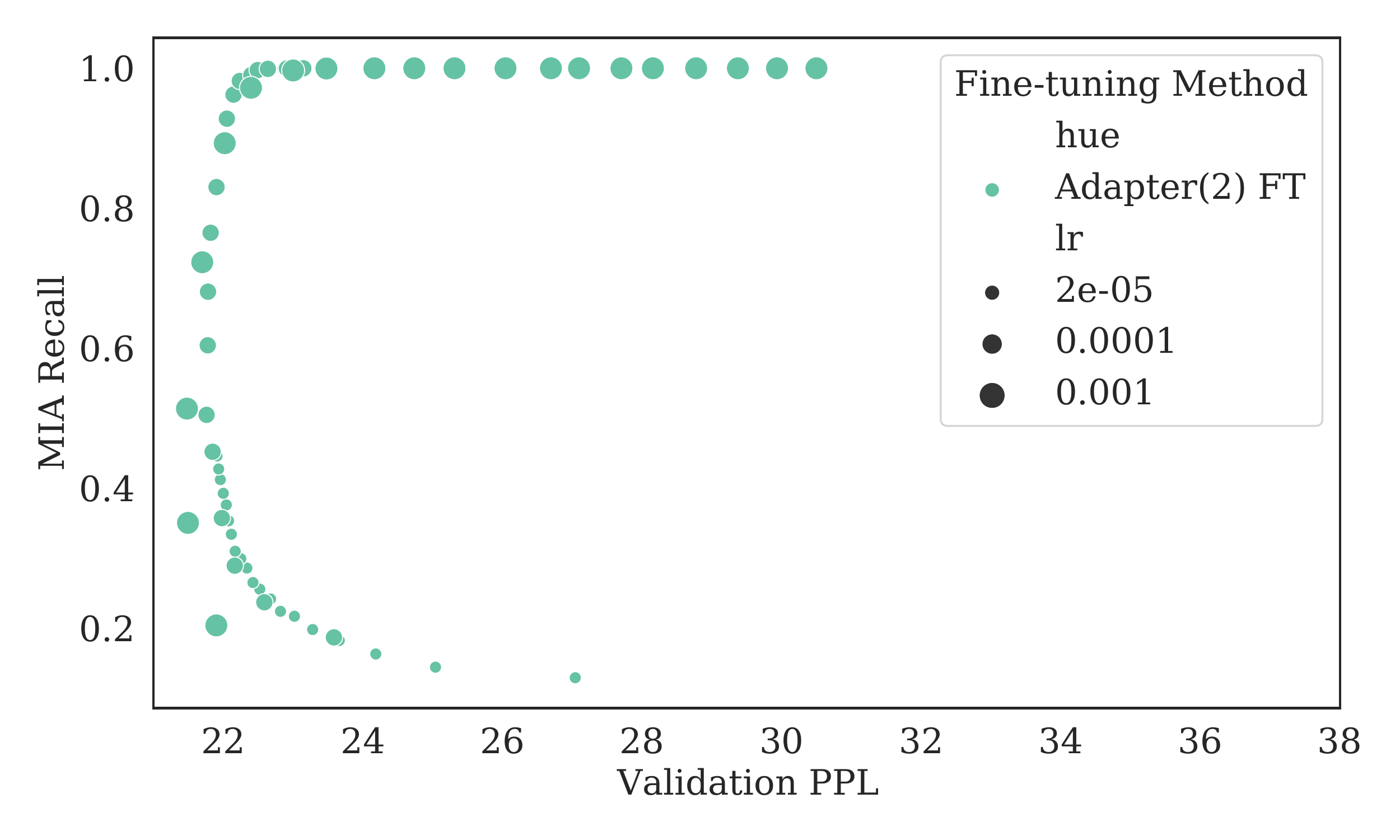}
    \footnotesize 
     \caption{Adapter(2) FT}
     \label{fig:sep:wikipedia:adp2}
    \end{subfigure}
  \caption{Wikipedia} 
    \label{fig:sep:wikipedia}
    \vspace{-2ex}
\end{figure}

\begin{figure}[]
    \centering
    
    \begin{subfigure}{0.48\textwidth}
     \includegraphics[width=\linewidth]{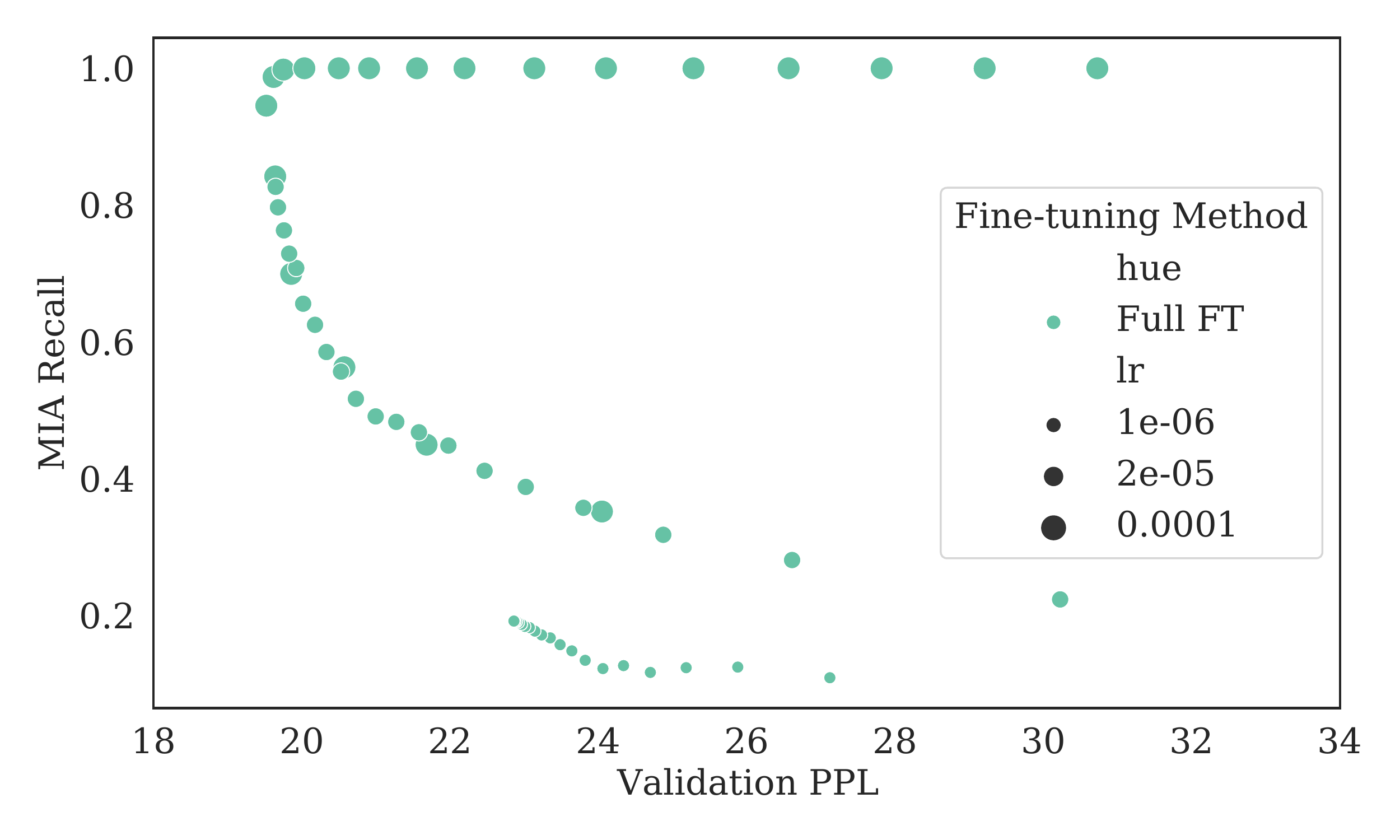}
     \footnotesize
     \caption{Full FT}
     \label{fig:sep:ptb:full}
    \end{subfigure}
    \begin{subfigure}{0.48\textwidth}
     \includegraphics[width=\linewidth]{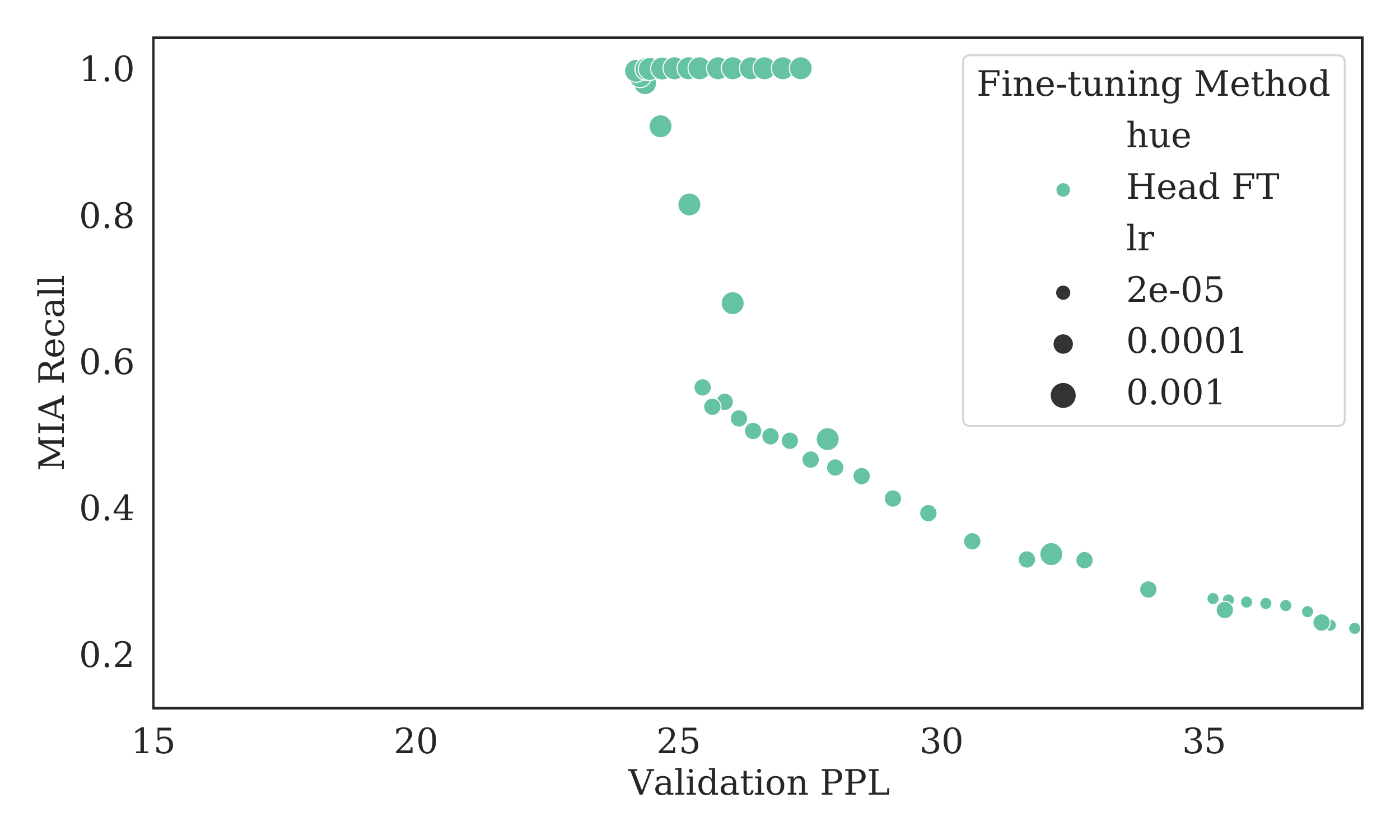}
    \footnotesize 
     \caption{Head FT}
     \label{fig:sep:ptb:headft}
    \end{subfigure}
    \begin{subfigure}{0.48\textwidth}
     \includegraphics[width=\linewidth]{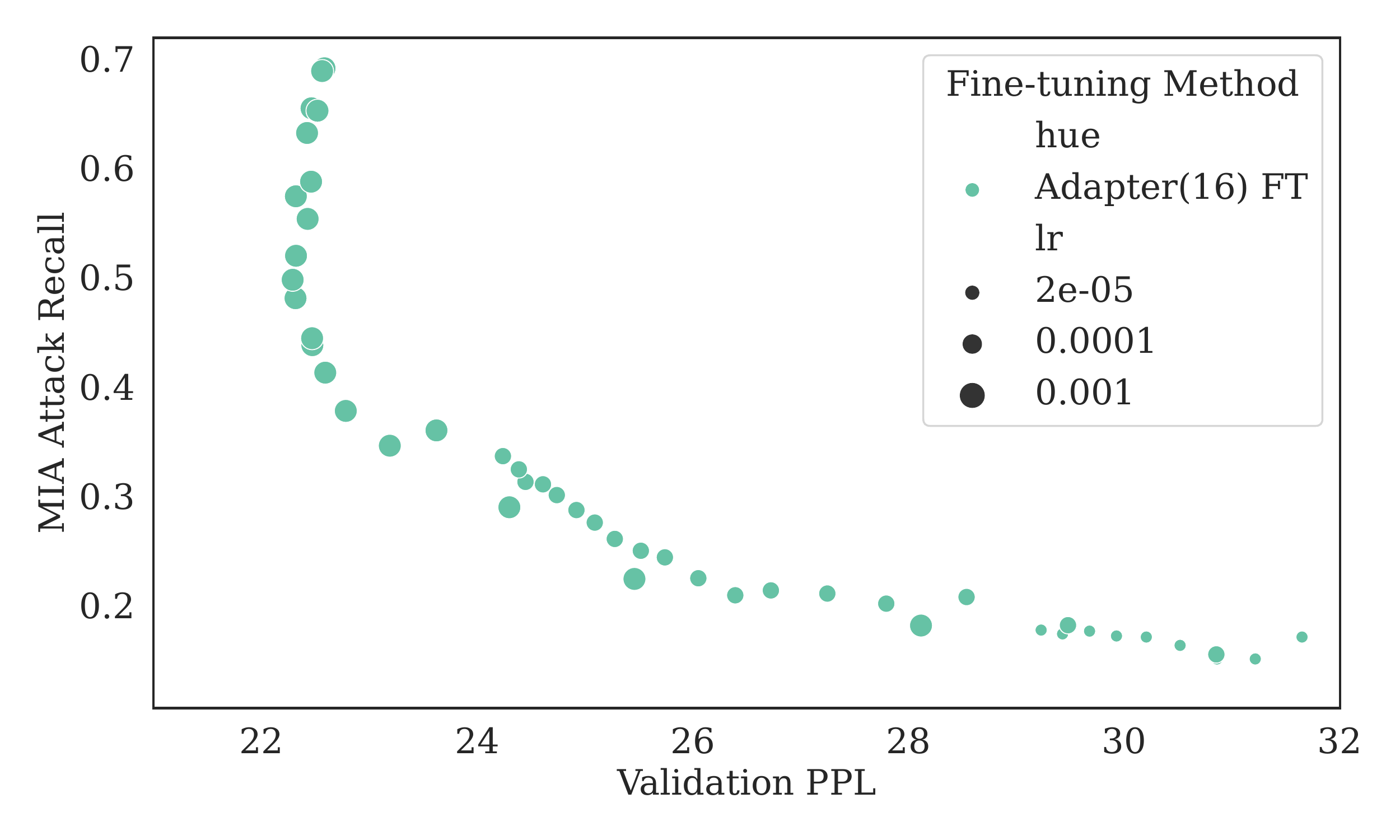}
    \footnotesize 
     \caption{Adapter(16) FT}
     \label{fig:sep:ptb:adp16}
    \end{subfigure}
    \begin{subfigure}{0.48\textwidth}
     \includegraphics[width=\linewidth]{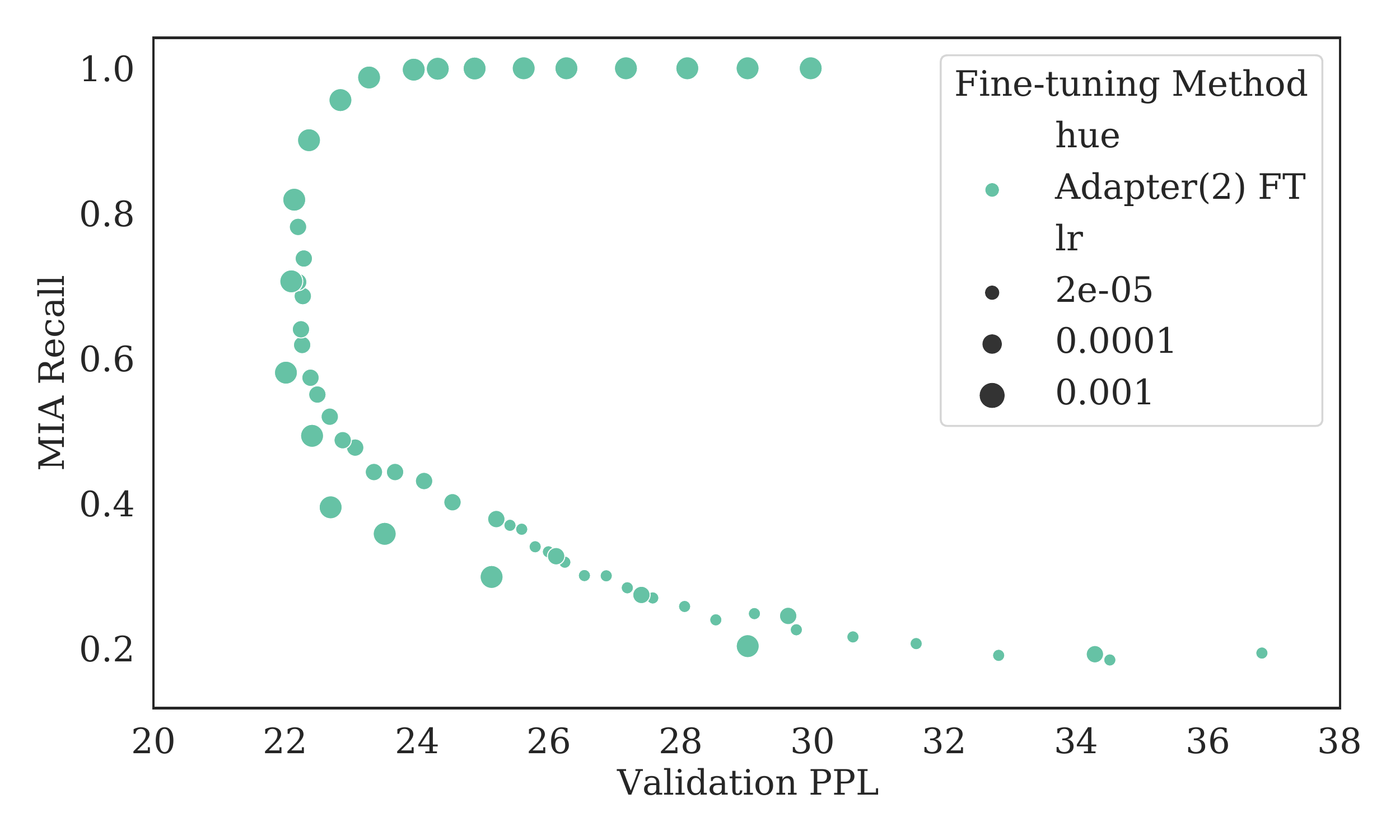}
    \footnotesize 
     \caption{Adapter(2) FT}
     \label{fig:sep:ptb:ad2}
    \end{subfigure}
  \caption{Penn Tree Bank} 
    \label{fig:sep:ptb}
    \vspace{-2ex}
\end{figure}

\begin{figure}[]
    \centering
    
    \begin{subfigure}{0.48\textwidth}
     \includegraphics[width=\linewidth]{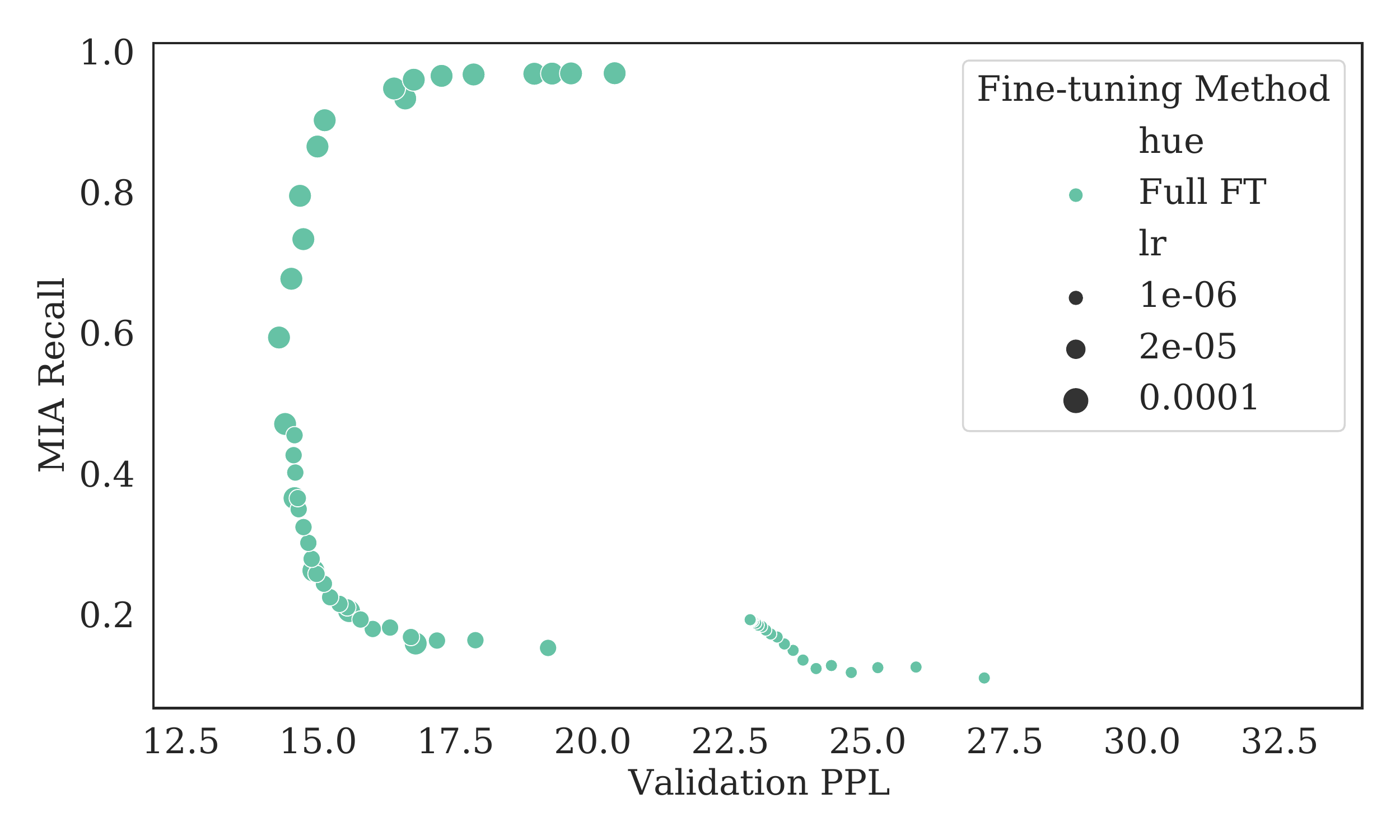}
     \footnotesize
     \caption{Full FT}
     \label{fig:sep:enron:full}
    \end{subfigure}
    \begin{subfigure}{0.48\textwidth}
     \includegraphics[width=\linewidth]{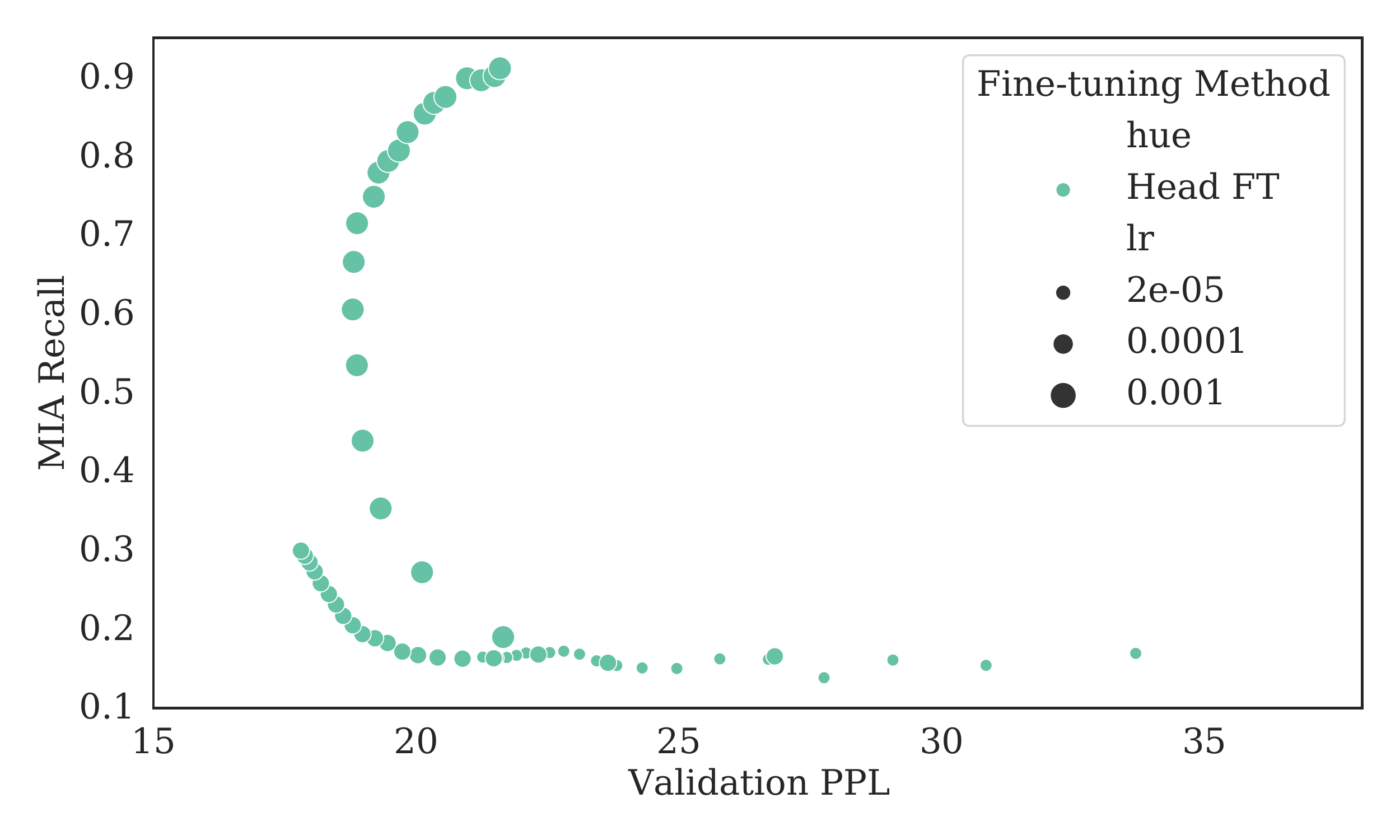}
    \footnotesize 
     \caption{Head FT}
     \label{fig:sep:enron:head}
    \end{subfigure}
    \begin{subfigure}{0.48\textwidth}
     \includegraphics[width=\linewidth]{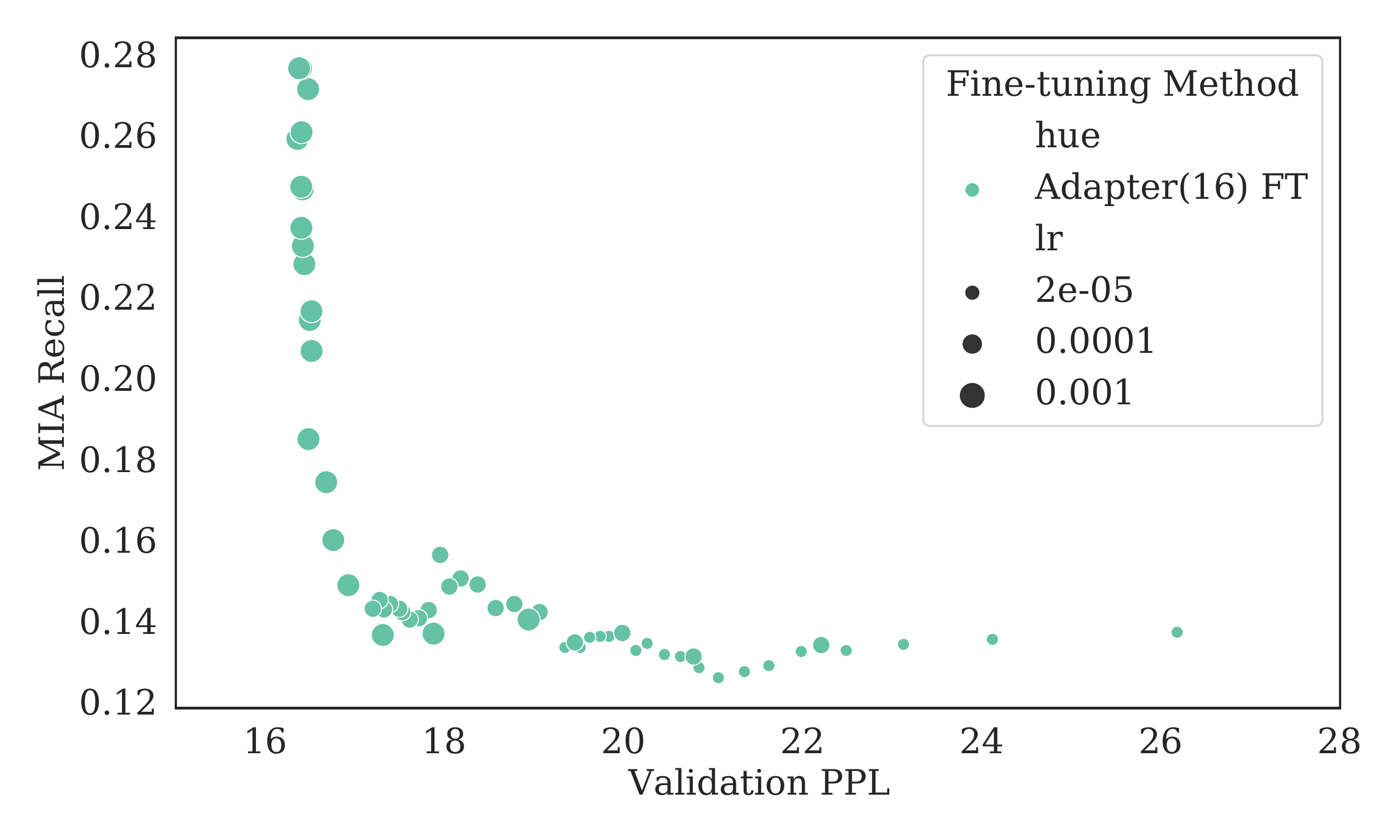}
    \footnotesize 
     \caption{Adapter(16) FT}
     \label{fig:sep:enron:adp16}
    \end{subfigure}
    \begin{subfigure}{0.48\textwidth}
     \includegraphics[width=\linewidth]{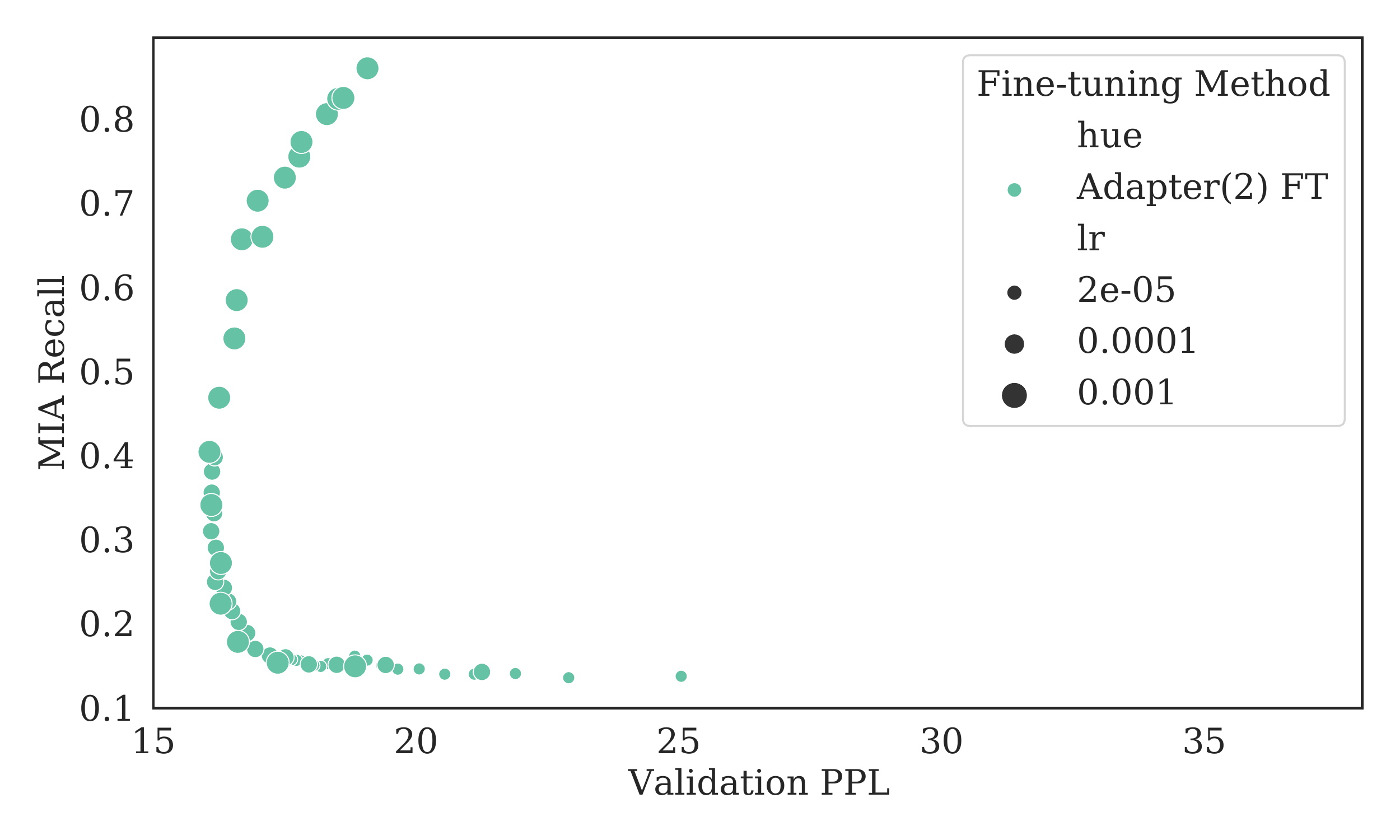}
    \footnotesize 
     \caption{Adapter(2) FT}
     \label{fig:sep:enron:adp2}
    \end{subfigure}
  \caption{Enron} 
    \label{fig:sep:enron}
    \vspace{-2ex}
\end{figure}

Figures~\ref{fig:sep:wikipedia},~\ref{fig:sep:ptb}, and~\ref{fig:sep:enron} show the MIA recall vs validation PPL for each fine-tuning method on each dataset separately, to provide better visibility. These Figures correspond to the subfigures in Figure~\ref{fig:pareto}.

\subsection{Breaking Fine-tuning Into Phases}
Although there is no ground truth rule on how the phases are defined, we use the following heuristic: break the training between phases 1 and 2 at points where the slope of the lines on the graph starts increasing drastically. For breaking between phases 2 and 3 we choose the point where the validation perplexity starts increasing again.

\subsection{Computational Resources}
For this paper, we spent an overall 7 days in GPU time for training and evaluation. For that, we used a server with 4$\times$RTX2080 GPU with 11GB of memory. 
%

\begin{figure}[]
    \centering
    \begin{subfigure}{0.48\textwidth}
     \includegraphics[width=\linewidth]{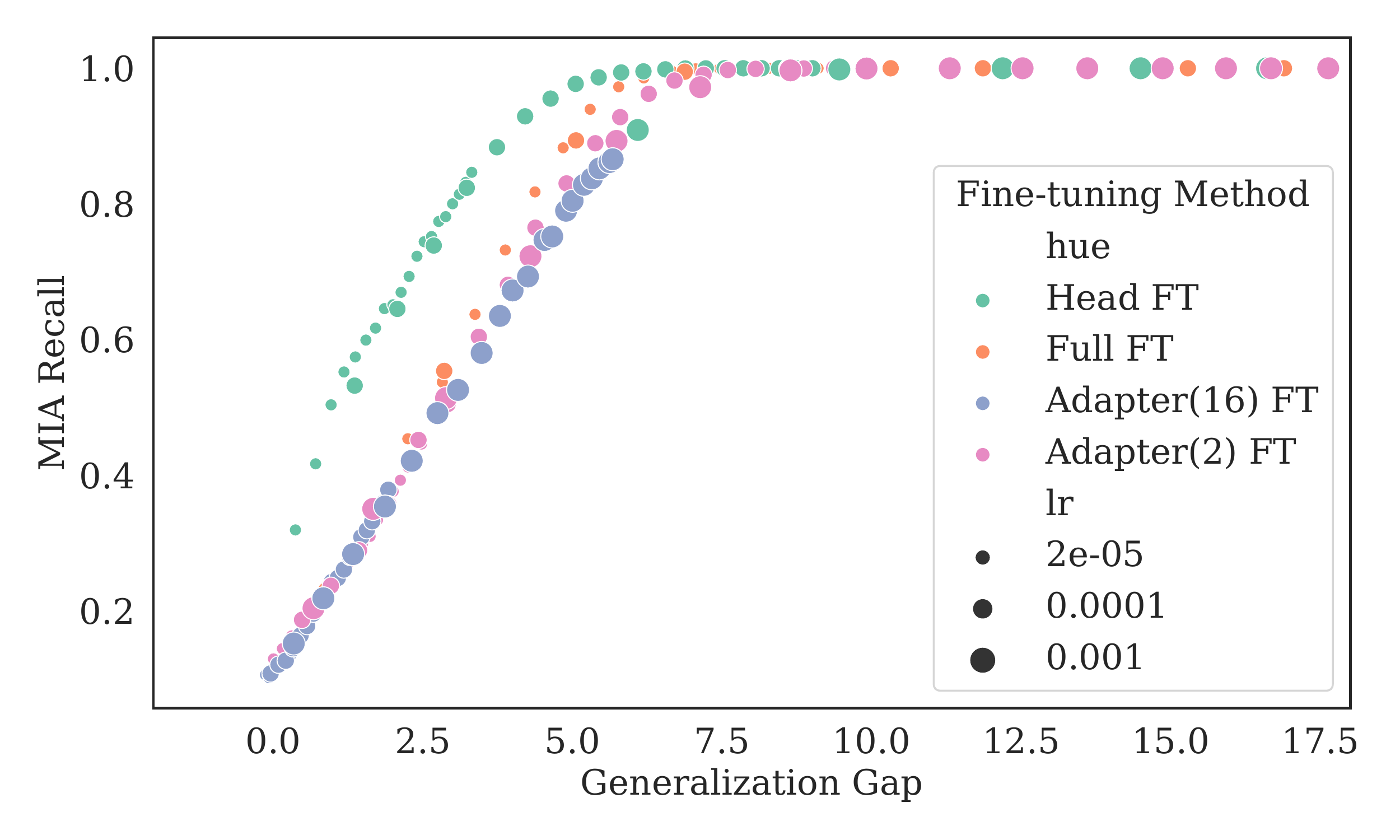}
     \footnotesize
     \caption{Wikipedia Dataset}
     \label{fig:gap:wikipedia}
    \end{subfigure}
    \begin{subfigure}{0.48\textwidth}
     \includegraphics[width=\linewidth]{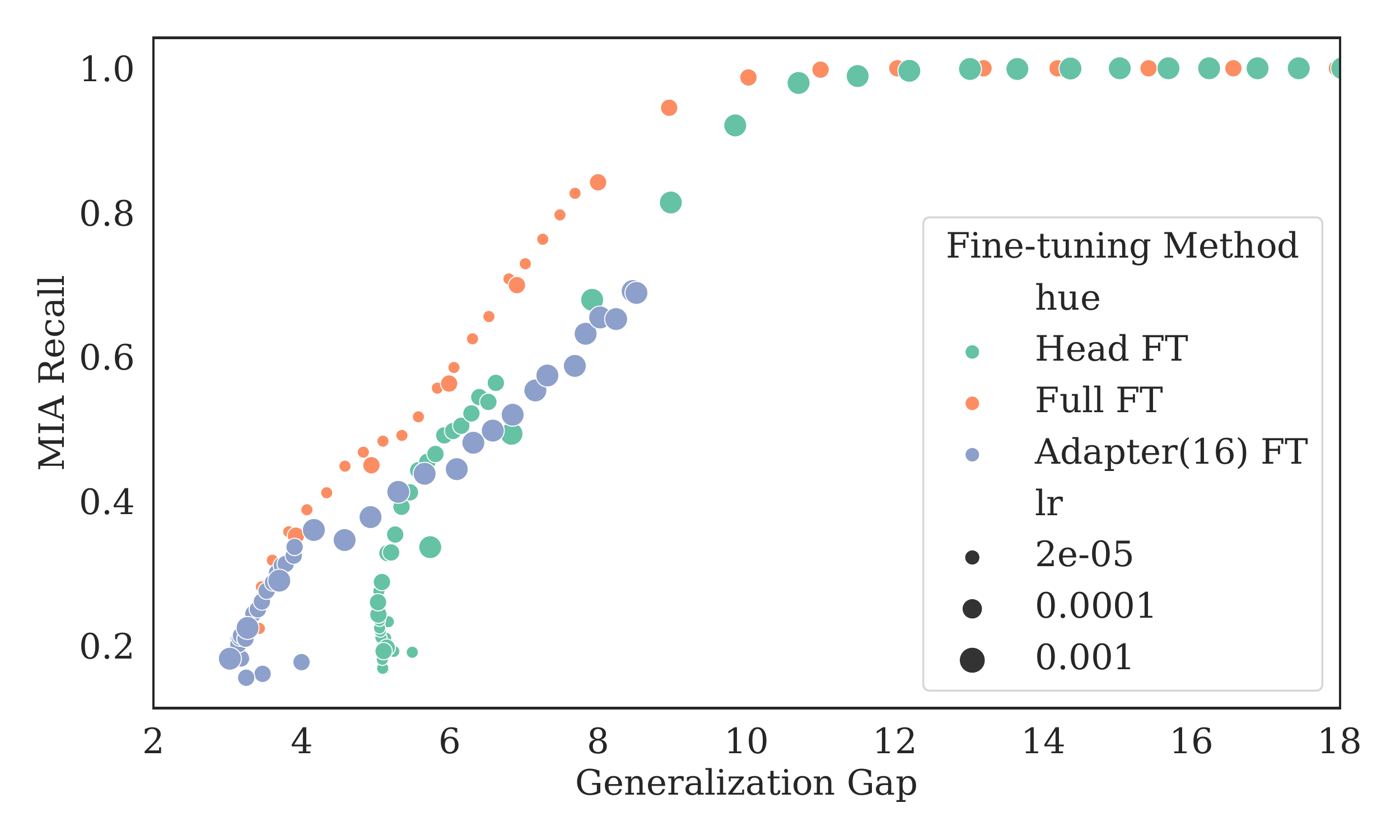}
    \footnotesize 
     \caption{Penn Treebank Dataset}
     \label{fig:gap:ptb}
    \end{subfigure}
    \begin{subfigure}{0.48\textwidth}
     \includegraphics[width=\linewidth]{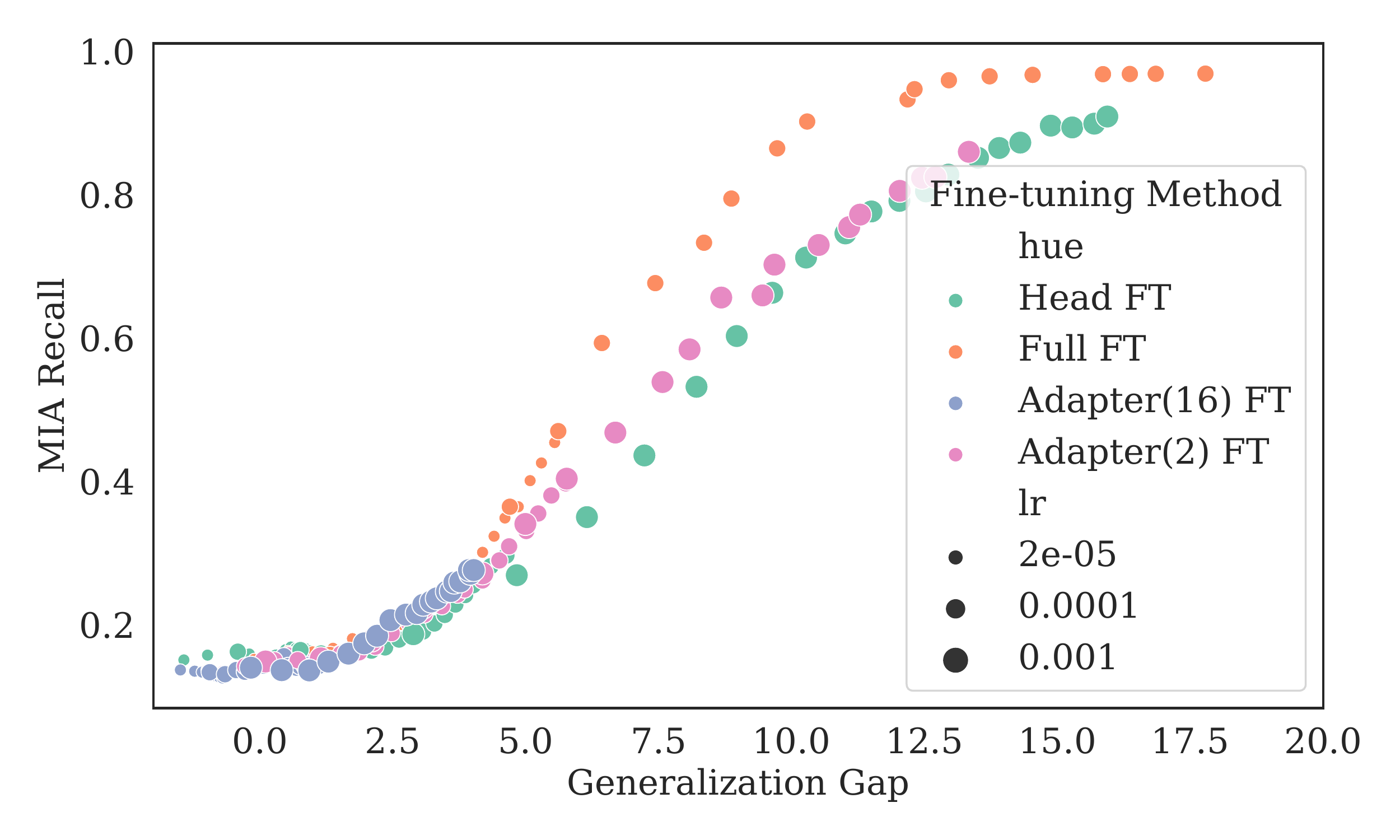}
    \footnotesize 
     \caption{Enron}
     \label{fig:gap:enron}
    \end{subfigure}
  \caption{Attack recall and generalization gap (Validation PPL- Train PPL) correlation. As the generalization gap increases, the attack observes more leakage as expected for all fine-tuning methods on both datasets. } 
    \label{fig:gap}
    \vspace{-2ex}
\end{figure}

\begin{figure}[]
    \centering
    \begin{subfigure}{0.48\textwidth}
     \includegraphics[width=\linewidth]{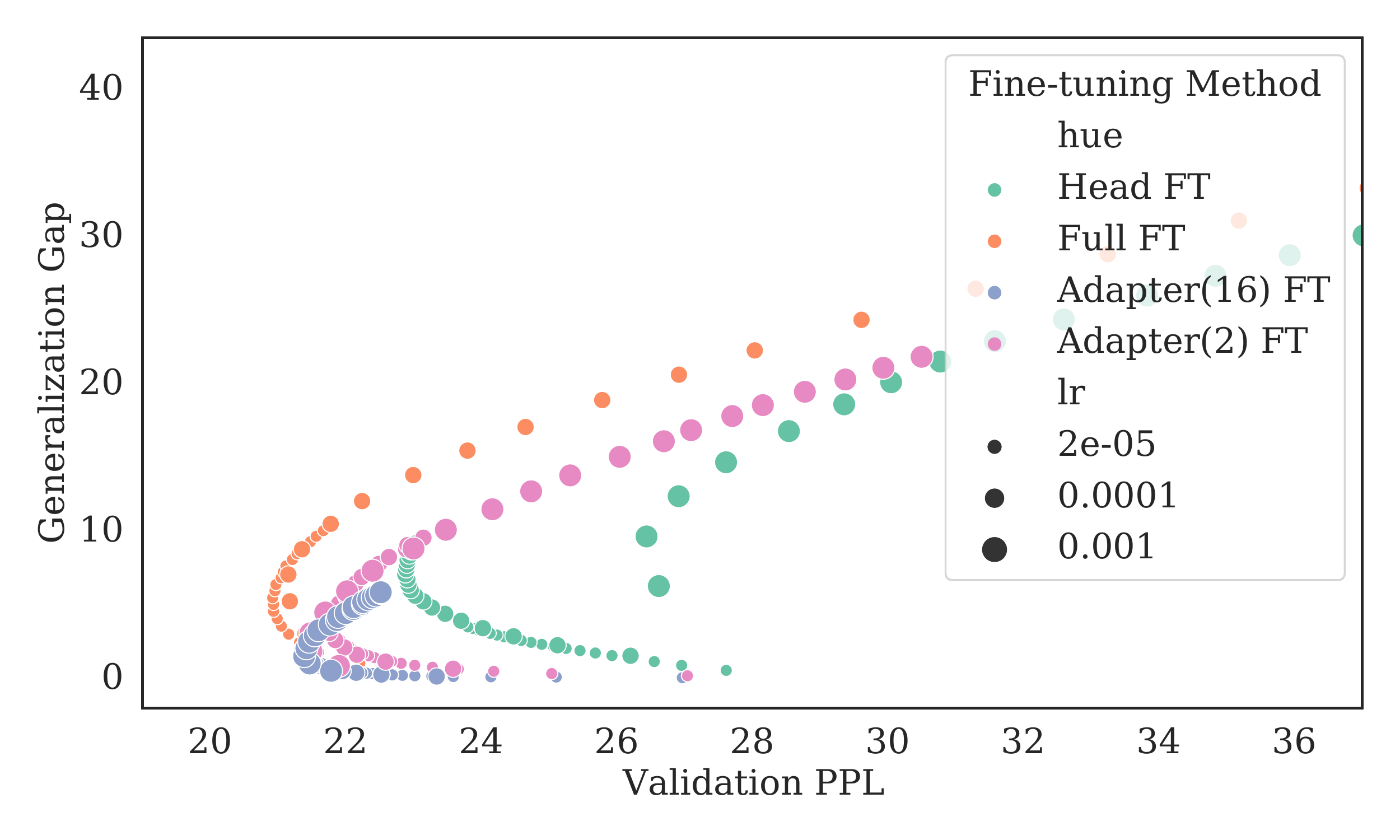}
     \footnotesize
     \caption{Wikipedia Dataset}
     \label{fig:gap:wikipedia}
    \end{subfigure}
    \begin{subfigure}{0.48\textwidth}
     \includegraphics[width=\linewidth]{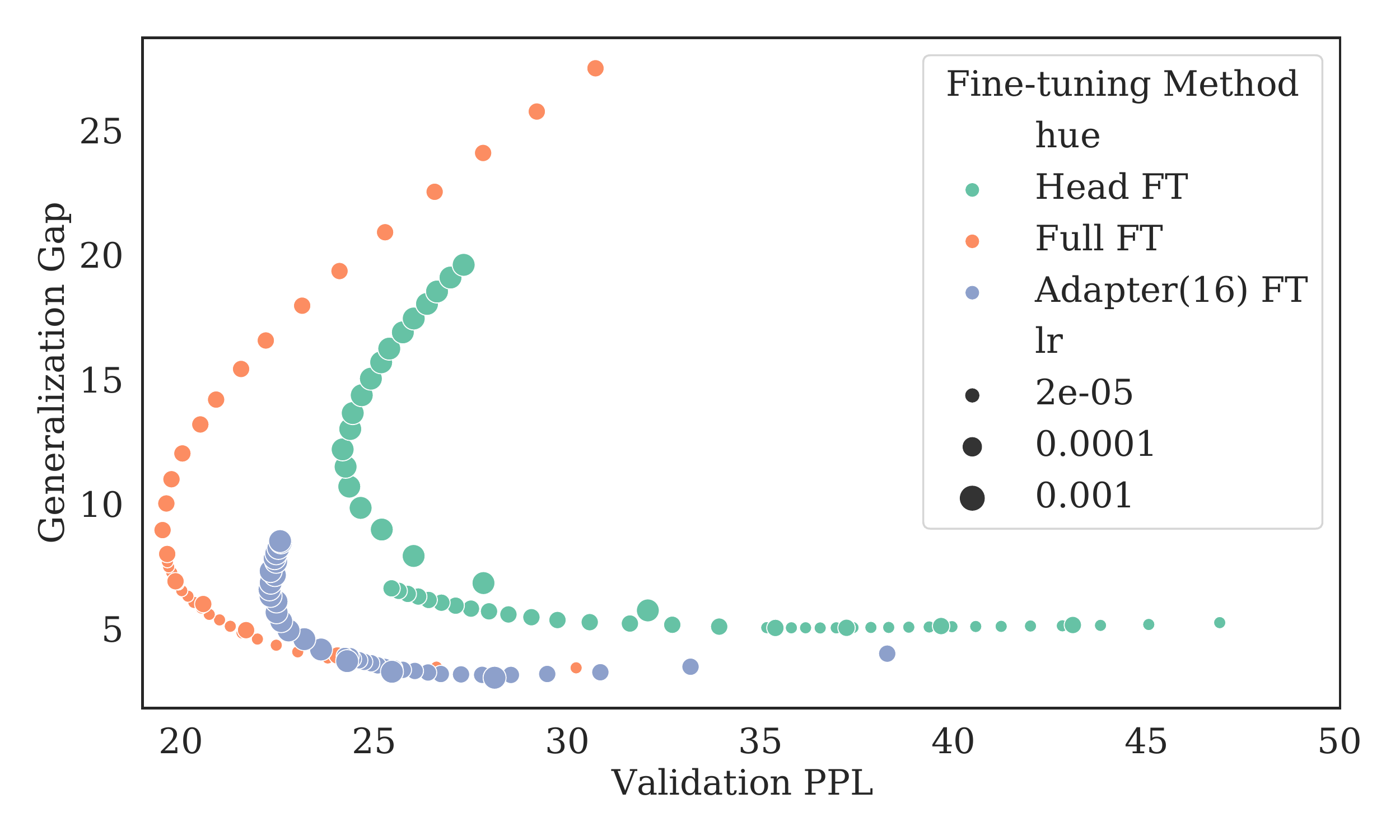}
    \footnotesize 
     \caption{Penn Treebank Dataset}
     \label{fig:gap:ptb}
    \end{subfigure}
    \begin{subfigure}{0.48\textwidth}
     \includegraphics[width=\linewidth]{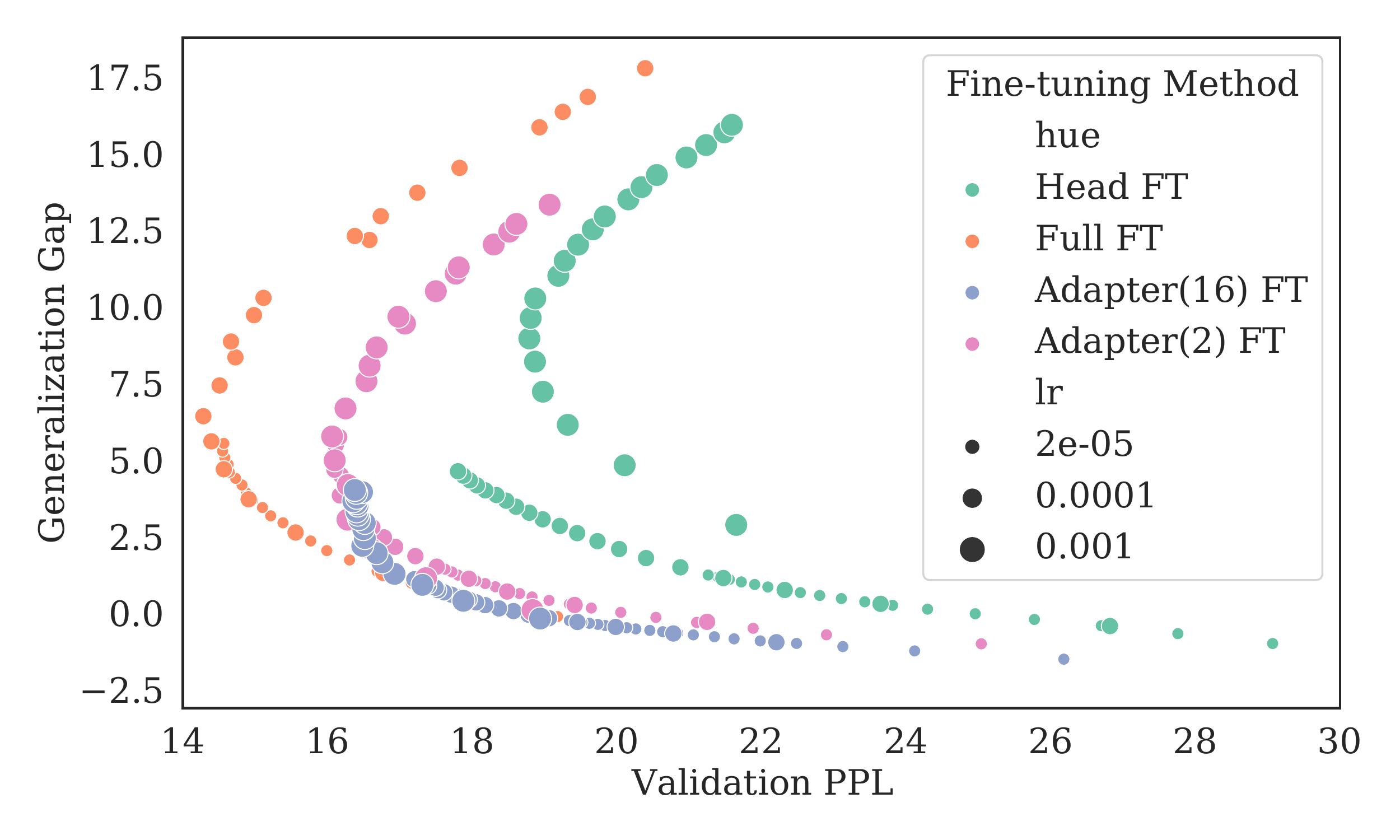}
    \footnotesize 
     \caption{Enron}
     \label{fig:gap:enron}
    \end{subfigure}
  \caption{Validation perplexity and generalization gap (Validation PPL- Train PPL) correlation. } 
    \label{fig:gap}
    \vspace{-2ex}
\end{figure}


\end{document}